\PassOptionsToPackage{table}{xcolor}

\documentclass{article}

\newif\ificlrstyle
\IfFileExists{iclr2027_conference.sty}{
  \iclrstyletrue
  \usepackage{iclr2027_conference,times}
}{
  \iclrstylefalse
  \usepackage[letterpaper,margin=1in]{geometry}
  \usepackage{natbib,times}
  \PackageWarningNoLine{LJ-WAM}{
    ICLR style missing; article syntax preview only
  }
}

\usepackage{
  amsmath,
  amssymb,
  mathtools,
  graphicx,
  booktabs,
  tabularx,
  array,
  makecell,
  multirow,
  xcolor
}

\usepackage{hyperref}
\usepackage{url}
\usepackage{epigraph}
\usepackage{wrapfig}
\usepackage{xspace}
\usepackage{caption}
\usepackage{subcaption} 
\usepackage{bbding}

\usepackage{pgfplots}
\pgfplotsset{compat=1.18}

\newcolumntype{Y}{>{\raggedright\arraybackslash}X}

\definecolor{oursblue}{RGB}{58,110,165}
\definecolor{ourslight}{RGB}{239,246,252}
\definecolor{baselinegray}{RGB}{112,164,174}
\definecolor{plotgray}{HTML}{A6A6A6}
\definecolor{plotblue}{HTML}{76B7D8}
\definecolor{plotgold}{HTML}{E3AA27}
\definecolor{plotgreen}{HTML}{5AA469}
\definecolor{plotdarkgreen}{HTML}{2F7F4F}
\newcommand{\best}[1]{\textbf{#1}}

\newcommand{\eg}{e.g.\@\xspace}

\newcommand{\aka}{a.k.a.\@\xspace}

\newcommand{\model}{\textsl{RoboActualizer}}

\newcommand{\MainOOD}{63.1}
\newcommand{\TrainParams}{60M}

\newcommand{\blfootnote}[1]{%
  \begingroup
  \renewcommand{\thefootnote}{}%
  \hypersetup{hidelinks}%
  \NoHyper
  \footnotetext{#1}%
  \endNoHyper
  \endgroup
}
\definecolor{darkred}{RGB}{210, 0, 0}

\ExplSyntaxOn
\seq_new:N \l__gray_row_cells_seq

\NewDocumentCommand{\grayrow}{m}{
  \seq_set_split:Nnn \l__gray_row_cells_seq { & } {#1}
  \seq_set_map:NNn
    \l__gray_row_cells_seq
    \l__gray_row_cells_seq
    {\textcolor{black!50}{##1}}
  \seq_use:Nn \l__gray_row_cells_seq { & }
}
\ExplSyntaxOff

\title{One from Infinity: Actualizing Futures from Pretrained World Models into Robot Actions}
\author{
Bang Du$^{*}$,
Yichen Xie$^{*,\dagger}$,
Shuqi Zhao$^{*}$,
Yuxin Chen,
Menglin Wu,
Masayoshi Tomizuka\\
~\\
\makebox[\textwidth][c]{University of California, Berkeley}\\
~\\
\makebox[\textwidth][c]{
\texttt{Project Webpage: \href{https://zhao-sq.github.io/RoboActualizer/}{\textcolor{darkred}{\model{}} }}}
}
\iclrfinalcopy
\begin{document}
\maketitle
\pagestyle{fancy}
\fancyhf{}
\fancyhead[L]{Preprint}
\renewcommand{\headrulewidth}{0.4pt}

\blfootnote{$^*$ Equal contribution}
\blfootnote{$^{\dagger}$ Correspondence: yichen\_xie{\char64}berkeley.edu}

\begin{abstract}
A pretrained video world model admits many plausible futures for a scene, but a robot must realize the exact task-conditioned one. To turn world models into executable robot policies, existing methods fine-tune the heavy world model backbone using large-scale robot data and computational resources. Challenging this \textit{status quo}, we argue that the expensive part has already been paid in the world model pretraining since the representation space of a video world model encoder lays out the diverse potential futures. In this case, what remains is to determine the future that accomplishes the task and to read out the actions that realize it. We formalize this task as \emph{actualization}, which learns a task-conditioned realization on top of a prior supplied by a frozen world model encoder. This can be solved by a tiny actualizer model. We implement \model{} with as few as 60M parameters on top of a frozen world model encoder. The actualizer is composed of two lightweight DiT experts that jointly predict future latents and actions by flow matching. The model can be trained entirely on a single GPU with 32 GB peak memory. With up to $100\times$ fewer trainable parameters than existing WAMs and VLAs, \model{} reaches great performance on simulation benchmarks including LIBERO, LIBERO-Plus, RoboTwin~2.0 and five tasks on two real-world platforms, with a low latency of 39\,ms that allows real-time control. Our code is available at
\href{https://zhao-sq.github.io/RoboActualizer/}{\textcolor{darkred}{project webpage}}.

\end{abstract}

\begin{figure}[h]
    \centering
\includegraphics[width=0.5\linewidth]{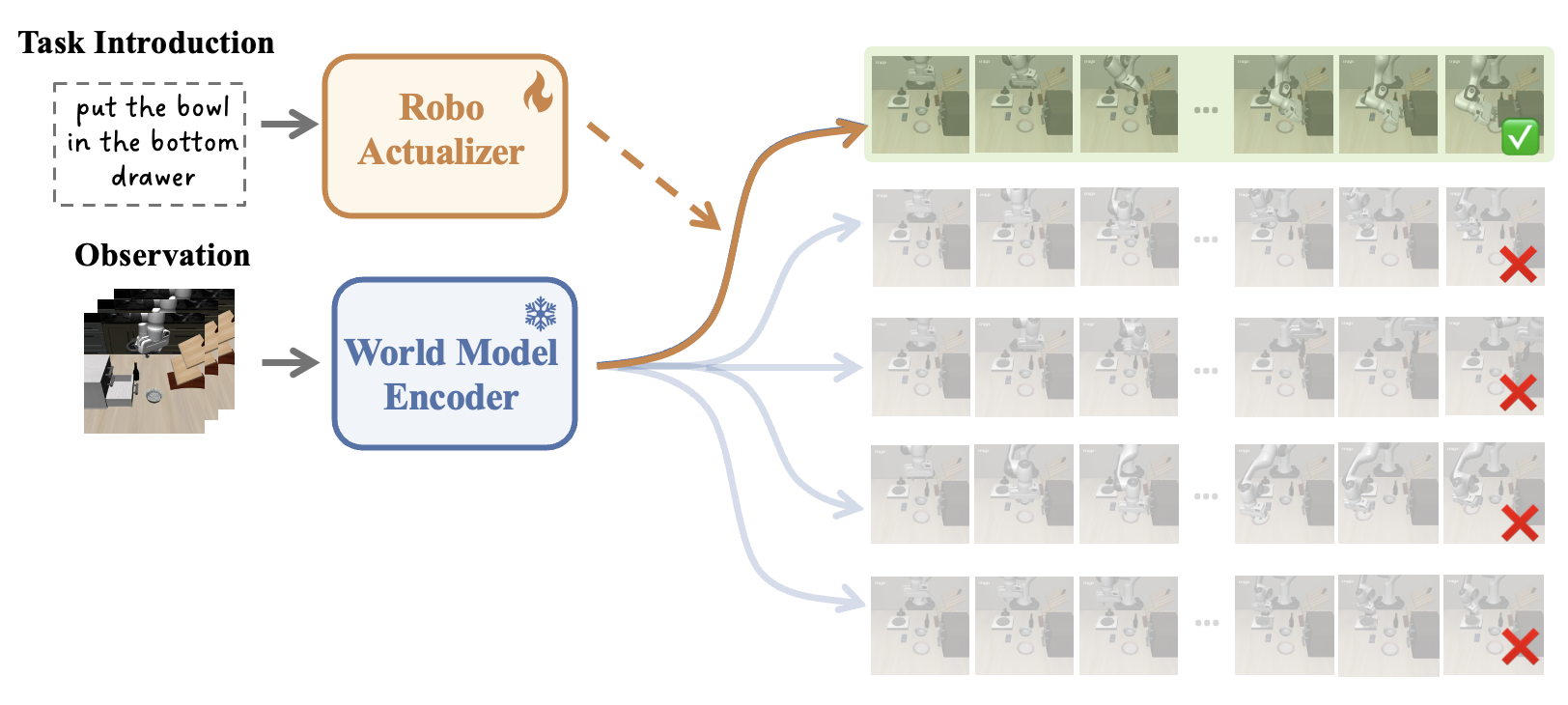}
\hfill
\includegraphics[width=0.23\linewidth]{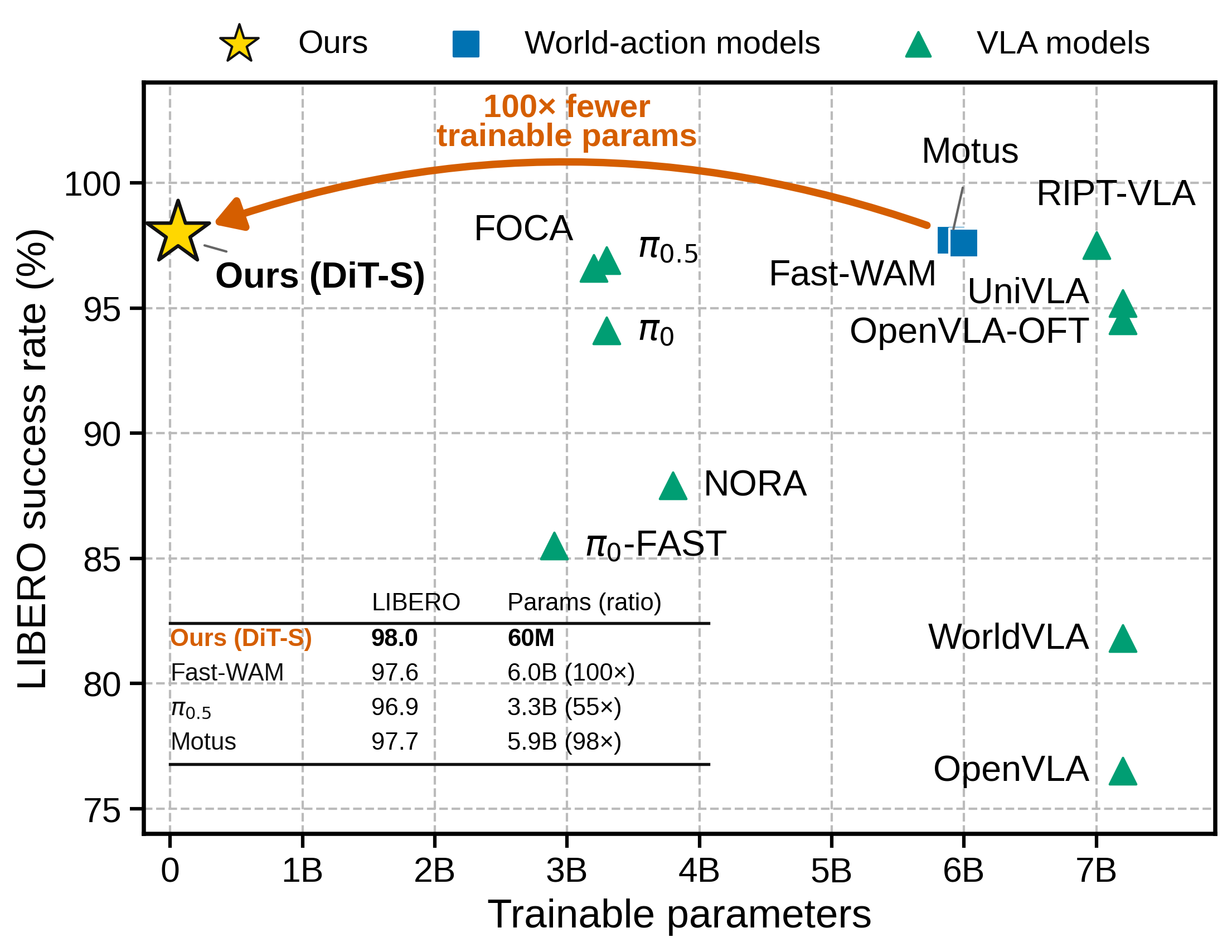}
\includegraphics[width=0.23\linewidth]{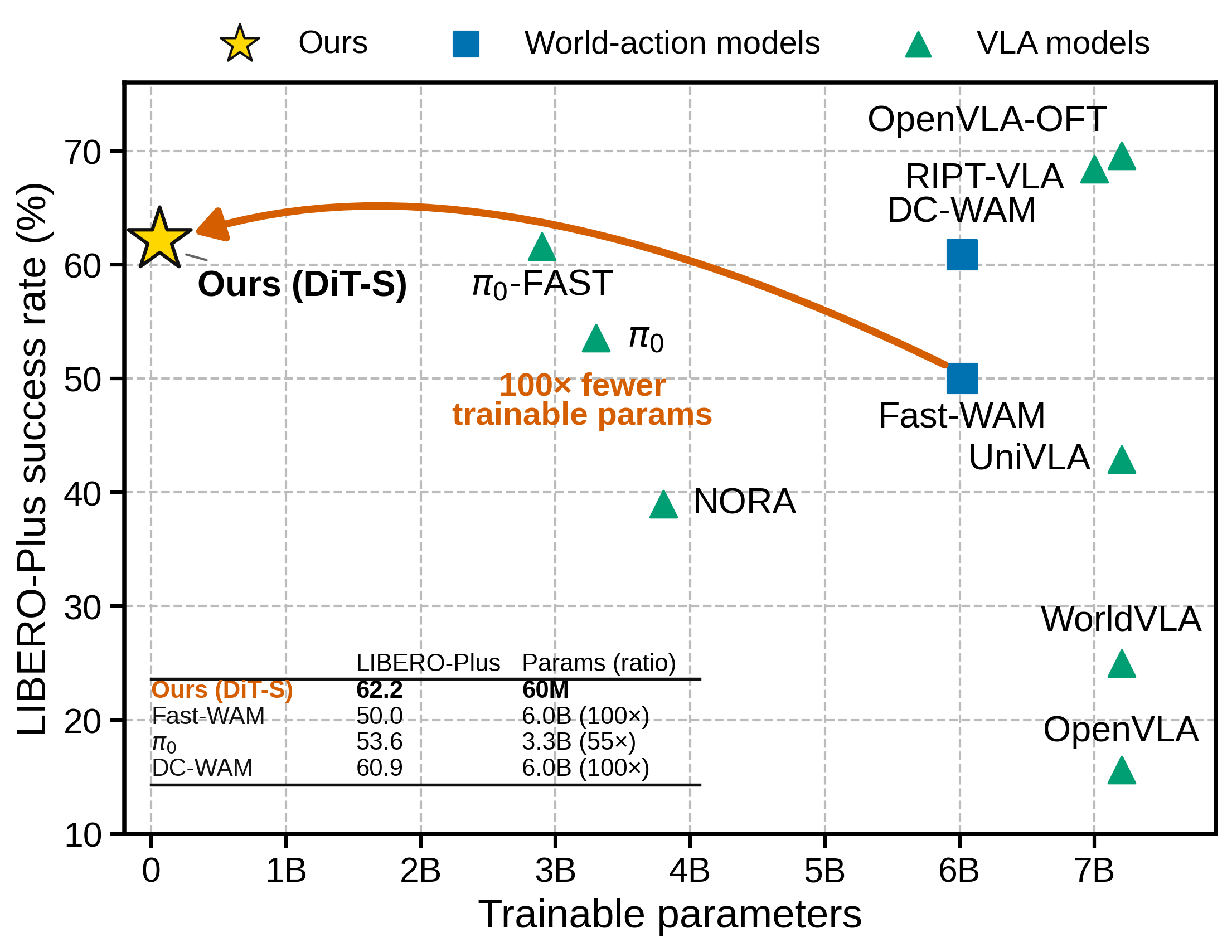}
        \caption{\model{} determines the task-conditioned future from the multiple plausible ones contained by the representation space (left). With as few as 60\,M parameters, \model{} outperforms baselines with $100\times$ more parameters on simulation benchmarks (right).}
    \label{fig:teaser}
\end{figure}

\section{Introduction}
\label{sec:introduction}

\setlength{\epigraphwidth}{0.6\textwidth}
\epigraph{\itshape There is an infinity of possible worlds, and only one can exist.}{---\ G.~W.~Leibniz, \emph{Monadology}}

Video world models pretrained on Internet-scale data have become strong predictors of how the world unfolds. From a few context frames, these models~\citep{wan2025wan,assran2025vjepa2,team2026advancing} learn dense spatio-temporal representations that anticipate future motion, interactions, and scene changes. A defining property of such models is that their predictions are not committed to a single outcome: the same environment admits many possible behaviors, and a general world model must represent all of these potential evolutions rather than a single one.

This property poses a dilemma for robotics, where a robot must ultimately complete one specific task with precise actions. Existing methods bridge this gap at considerable expense. \emph{Planning-based} approaches keep the world model frozen and search over candidate actions at test time, typically with model-predictive control~\citep{zhou2024dino,maes2026leworldmodel}. Each decision then requires many imagined rollouts, which caps the control frequency, and the objective is usually a goal image or a hand-designed cost. Alternatively, \emph{world-action models} (WAMs) fine-tune a general foundation model so that action prediction is absorbed into the world model itself. This, however, requires updating a heavy backbone with substantial robot data and considerable computation~\citep{bjorck2025gr00t,li2026causal,ye2026world,yan2026flexpimultistreamworldactionmodel}. Either way, the conversion is costly. Given that enormous computation has already been invested in pretraining these visual foundation models, it is desired that the cost of converting \emph{future predictions} into \emph{robot actions} can be reduced.

In this paper, we seek \emph{an economical way to actualize diverse future predictions into robot actions}. Our key insight is that the latent space of a pretrained world model encodes the dynamics and possible futures of the scene, so what remains for the robot is not to learn the evolution again, but to \emph{determine} the one that accomplishes the task among those futures and to read out the actions that realize it (Fig.~\ref{fig:teaser}). This is a far cheaper problem than dynamics modeling, and we find that \emph{a lightweight module} suffices for it. The world model itself is never updated. All robot-related learning is confined to a tiny flow matching head. Generic pretraining remains an upstream investment, but it is paid once and amortized across tasks, whereas the downstream cost of obtaining a robot policy is reduced to training a tiny model on cached features.

Concretely, we propose \model{}, which pairs a frozen V-JEPA~2.1 world model encoder~\citep{mur2026v} with two compact diffusion transformers~\citep{peebles2023scalable} organized as a Mixture-of-Transformers (MoT) architecture~\citep{liang2024mixture}. The \emph{latent expert} predicts the future representation of the scene conditioned on the task instruction where the instruction thus acts as the condition to determine one from many future contained in the world model latent space, in which the task is accomplished. The future latent provides training-time supervision that teaches the model how to read the frozen representation. The \emph{action expert} predicts, from the same context, the continuous control signals that realize this future. Both are trained jointly from task demonstrations.  At deployment, the encoder runs a single forward pass per observation, and only the lightweight head runs repeatedly during denoising. This design gives \model{} several key advantages:
\begin{itemize}
    \item \textbf{Strong Performance.} \model{} achieves great performance on competitive simulation benchmarks, including LIBERO~\citep{liu2023libero} and RoboTwin~2.0~\citep{chen2025robotwin}, as well as five tasks on two different real-world robots. It consistently outperforms strong baselines despite only 60M trainable parameters (up to 100$\times$ fewer than baselines).
    \item \textbf{Generalization.} On LIBERO-Plus~\citep{fei2025liberoplus}, a perturbed variant of LIBERO, \model{} reaches \MainOOD\%, compared with 50.0\% for Fast-WAM~\citep{yuan2026fast} (6.0B trainable parameters). Exploiting the strong representations of the pretrained world model appears to preserve robustness that fine-tuning tends to erode.
    \item \textbf{Low Cost.} \model{} has approximately \TrainParams{} trainable parameters and is optimized on cached latents together, these allowing the entire model to be trained on \emph{a single GPU} without any embodied pretraining. This substantially lowers the barrier to training world-action models.
    \item \textbf{Efficiency.} Because denoising occurs only within the small actualizer, the deployed policy runs at more than \emph{25\,Hz} on a NVIDIA RTX A6000 GPU, satisfying the requirements of real-time robot control.

\end{itemize}

\section{Related Work}
\label{sec:related_work}
\noindent\textbf{Video World Models.}
Generalizable world models are pretrained on Internet-scale video along two lines. The first generates future frames in pixel space, scaling video generation models into general-purpose simulators~\citep{openai2024sora,agarwal2025cosmos,wan2025wan,team2026advancing}. The second predicts the future in representation space. The V-JEPA family~\citep{bardes2024revisiting,assran2025vjepa2} learns representations by predicting masked spatio-temporal latents. DINO-WM~\citep{zhou2024dino} and NWM~\citep{bar2025navigation} model dynamics on top of frozen visual features~\citep{oquab2023dinov2}. Because latent prediction sidesteps the cost of synthesizing pixels, it yields high-level semantic features well suited to downstream robotic tasks. Both lines share the property we build on: a world model trained across many environments is not committed to a single continuation, and its predictions span the space of plausible futures. We take a pretrained latent world model as a fixed representation of this space and ask how cheaply a robot policy can be extracted from it.

\noindent\textbf{From World Models to Robot Actions.}
The predictive ability of world models is an appealing property for robot policies, and existing efforts extract actions from pretrained world models in two ways. \emph{Model predictive control (MPC)} keeps the world model fixed. For each imagined rollout, these methods generate future videos \citep{du2023learning} or latent~\citep{zhou2024dino,maes2026leworldmodel} and optimize actions against a task-specific reward. Planning preserves the pretrained model but pays the price at inference. Each action requires many rollouts, and the objective is typically a goal image or a hand-designed cost rather than a language instruction. The other approach \emph{fine-tunes} pretrained visual world models into world-action models (WAMs), additionally supervising the policy with future-frame generation in either cascaded or unified form~\citep{cheang2024gr,hu2024video,bjorck2025gr00t,ye2026world,yan2026flexpimultistreamworldactionmodel}. A rapidly growing line keeps future prediction as supervision but moves it from pixels into representation space. FLARE~\citep{zheng2025flare} aligns policy features with future latent embeddings. VLA-JEPA~\citep{sun2026vla} and FOCA~\citep{nguyen2026foca} supervises a VLA with a separate latent world model. Concurrent with our work, ST-WAM~\citep{wang2026stwam} adds DINOv3-space~\citep{simeoni2025dinov3} future experts to a video-generative WAM. JEPA-WAM~\citep{lin2026jepa} couples latent transition prediction and action generation through a shared language-model predictor on top of a frozen encoder. Despite building on V-JEPA~2.1~\citep{mur2026v}, JEPA-WAM encodes each observation as a single image and does not condition on an observation history. These works are motivated differently from ours. They treat future image latent as a better supervision target than pixels, but the observation itself is still encoded frame by frame, so the features carry no cues about how the scene is evolving, and all temporal reasoning must be learned by a heavy trainable model. In contrast, \model{} encodes a short observation history, so that motion and plausible continuations are already present in the features. What remains is to actualize one of these futures for the task, an easier goal that a tiny trained module suffices for.

\noindent\textbf{Pretrained Visual Representations for Robot Control.}
A parallel line of work applies general visual representations pretrained on large-scale datasets to robot control without task-specific fine-tuning~\citep{parisi2022unsurprising,goodwin2022zero,xia2025cage,xie2026multi}, with encouraging results. Others go a step further and learn robot-oriented representations tailored to downstream tasks~\citep{nair2022r3m,shang2024theia,wang2026ver}. However, frozen pretrained features are sensitive to domain mismatch.
Our results suggest a different interpretation: the gap reflects how the representation is read out rather than what it contains. V-JEPA~2.1~\citep{mur2026v} provides dense features that encode diverse possible scene evolutions, and \model{} uses them both as policy inputs and as future-prediction targets to learn a lightweight actualizer that determines the task-conditioned future. 
In this sense, \model{} shows that freezing is compatible with strong control when the readout is designed accordingly.

\section{Methodology}
\label{sec:method}
\begin{figure}[t]
  \centering
  \includegraphics[width=0.85\linewidth]{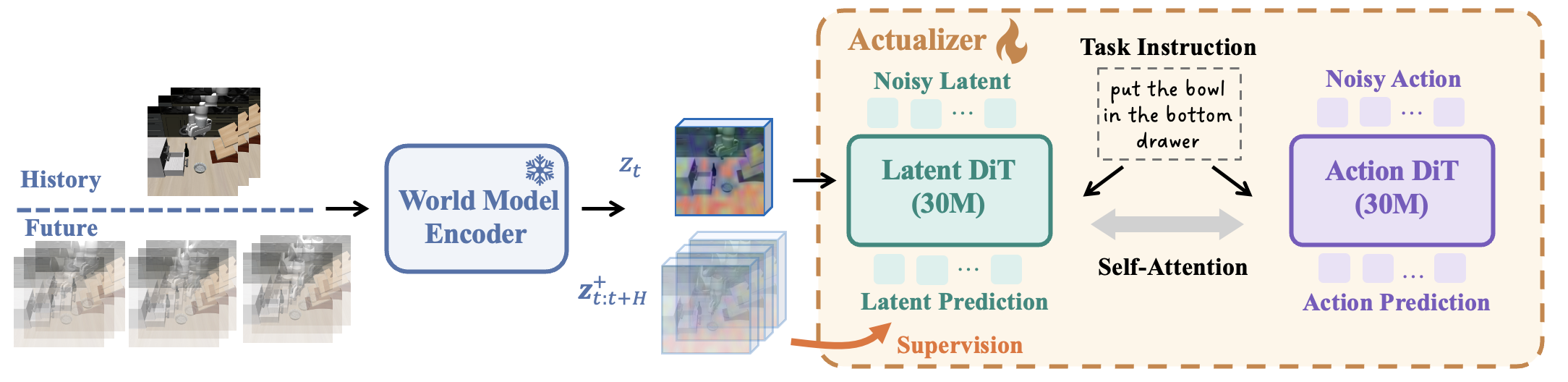}
  \vspace{-10pt}
  \caption{Overview of \model{} framework. \model{} is built upon a frozen world model encoder. It actualizes the task-conditioned latent future with a lightweight model.}
  \label{fig:architecture}
  \vspace{-10pt}
\end{figure}

This section explains the design of \model{} (Fig.~\ref{fig:architecture}). We formalize actualization as determine one future from the many contained in a pretrained world model latent space and realizing it in actions (Sec.~\ref{sec:formulation}), then instantiate the world model encoder with a frozen V-JEPA~2.1 (Sec.~\ref{sec:representation}). Sec.~\ref{sec:why_small} argues that a narrow actualizer is enough, and Sec.~\ref{sec:actualizer} presents the detailed implementation of actualizer model.

\subsection{Formulation: Actualizing One Future from Many}
\label{sec:formulation}

Our method rests on a division of labor between a pretrained world model encoder and a small learned actualizer. The world model encoder contains many plausible future evolutions of a scene, while the instruction picks one of them and produces the actions that realize it. 

\noindent\textbf{What the world model encoder provides.} The pretrained world model encoder provides a representation of the scene that already contains its future dynamics and evolutions. A video world model encoder $g_{\phi}$, pretrained on large-scale generic video, encodes a short clip of RGB observations into a latent representation
\begin{equation}
    z_t= g_{\phi}(\mathbf{x}_{t-\Delta:t})\in\mathbb R^{N\times D},
    \label{eq:latent}
\end{equation}
where the clip $\mathbf{x}_{t-\Delta:t}$ spans $\Delta$ control steps ending at step $t$, and $z_t$ consists of $N$ spatial tokens of dimension $D$. Because $g_{\phi}$ attends across frames, $z_t$ encodes the clip as a whole rather than frame by frame. This representation already captures the motion in the window and indicates what can plausibly happen next. The dynamics of the scene are thus built into the representation space, so a downstream learner does not have to acquire them from scratch by itself.

\noindent\textbf{What the world model encoder leaves undetermined.} The world model encoder does not determine which of these futures will occur. Knowing what \emph{can} happen is not knowing what \emph{will} happen since the same initial scene can be paired with multiple different tasks (Fig.~\ref{fig:future_latent_alignment}). Let $\mathcal{L}$ be the set of task instructions and $\mathbf{z}^+_{t:t+H}$ be the latent futures over the next $H$ steps. The futures that may follow $z_t$ form a mixture distribution with one component corresponding to each task:
\begin{equation}
  p(\mathbf{z}^+_{t:t+H}\mid z_t)\;=\;\sum_{\ell\in{\mathcal{L}}}\,p(\ell\mid z_t)\;p(\mathbf{z}^+_{t:t+H}\mid z_t,\ell).
  \label{eq:mixture}
\end{equation}
Each component is concentrated on the futures in which task $\ell$ is accomplished. We call the region it occupies $\mathcal F_\ell(z_t)$, then the union $\mathcal F(z_t)=\bigcup_\ell\mathcal F_\ell(z_t)$ is everything the scene can become. Given the latent representation $z_t$ alone, the world model encoder admits all of $\mathcal F(z_t)$ in its latent space but says nothing about which component will be realized.

\noindent\textbf{What actualizer remains to learn.} The actualizer learns the task-conditioned admission among these futures and the actions that carry it out. Conditioned on the instruction $\ell$, the joint distribution of future and actions factors into two:
\begin{equation}
q_\theta(\mathbf z^+_{t:t+H},\mathbf a_{t:t+H}\mid z_t,\ell)=
\underbrace{q_\theta(\mathbf z^+_{t:t+H}\mid h_t,\ell)}_{\text{\textbf{determination}: fits } p(\mathbf z^+_{t:t+H}\mid z_t,\ell)\text{in Eq. (2)}}
\cdot
\underbrace{q_\theta(\mathbf a_{t:t+H}\mid h_t,\ell)}_{\text{\textbf{realization:} action prediction}},\quad h_t=f_\theta(z_t,\ell)
  \label{eq:factors}
\end{equation}
where a task-conditioned scene state $h_t$ that summarizes the determined future.
From left to right, Eq.~\eqref{eq:factors} fits one task-conditioned component from the prior mixture and predicts the corresponding actions.
The prior mixture in Eq.~\eqref{eq:mixture} is provided by the world model encoder. The latent space of $g_\phi$ contains the cues of such futures, so the region $\mathcal F(z_t)$ and its components are already indicated in the frozen space, and the actualizer inherits their geometry rather than learns it. The actualization reweights the prior toward the component of task $\ell$, and it is the only thing about the future that must be learned from robot data. The realization maps the determined future to the actions that produce it. We call learning the determination and the realization, with the prior held fixed, as \emph{actualization}.

Given robot demonstrations $\mathcal D$ of observations, instructions, and actions, the actualizer $q_\theta$ is fit to Eq.~\eqref{eq:factors} by maximum likelihood,
\begin{equation}
  \min_\theta\;\mathbb E_{\mathcal D}\Big[-\log q_\theta\big(\mathbf z^+_{t:t+H},\,\mathbf a_{t:t+H}\,\big|\,z_t,\ell\big)\Big],
  \qquad \text{s.t.}\;\; z_t=g_\phi(\mathbf x_{t-\Delta:t})\ \text{with}\ \phi\ \text{fixed},
  \label{eq:actualization}
\end{equation}

It is implemented as a lightweight flow-matching model in Sec.~\ref{sec:actualizer}. Although Eq.~\eqref{eq:actualization} looks similar with an ordinary policy objective, Eq.~\eqref{eq:factors} shows why it is much easier. The targets $\mathbf z^+_{t:t+H}$ are points in a frozen space and the prior is already encoded in $\mathbf z_t$, so for each instruction, $q_\theta$ only has to find the right component $\mathcal{F}_\ell(z_t)$ instead of building the components themselves.

In contrast, previous works fine-tune to fit the same posterior with $\phi$ among the optimized variables. The geometry of $\mathcal F$ is then no longer given, and the learner must shape the prior and the policy at once, which yields a far harder problem. The complexity of actualization, by contrast, is set by the future prediction ability of world model latent space rather than by the dynamics of the scene, which is why a tiny $q_{\theta}$ suffices.

\subsection{Future Priors from World Model Encoder}
\label{sec:representation}

We instantiate $g_\phi$ with frozen V-JEPA~2.1~\citep{mur2026v}, a video foundation model. As a world model encoder, the training objective of V-JEPA is to predict masked spatio-temporal latent from context, whose representation space naturally indicates the future prior, as described in Eq.~\eqref{eq:mixture}. 

The present is encoded as a clip. $z_t$ is computed from a short window of frames $\mathbf{x}_{t-\Delta:t}$ ending at $t$, where we extract the last frame latent as $z_t$ (Eq.~\eqref{eq:latent}). Thus, scene evolution priors enter the representation through the encoder's cross-frame modeling rather than being inferred downstream. Then, the futures are encoded by the same $g_\phi$ and in the same manner. 

The future targets $\mathbf z^+_{t:t+H}$ are discretized at a few time steps spread over the action chunk,
\begin{equation}
    \mathbf z^+_{t:t+H}=(z^+_{t,h})_{h\in\mathcal H},\qquad z^+_{t,h}=g_\phi(\mathbf x_{t+h-\Delta:t+h}).
\end{equation}
Since $g_\phi$ is frozen in the downstream policy learning, all the latents can be computed offline. We train the actualizer $q_\theta$ entirely on cached latents without a single-time forward pass of the world model encoder, which significantly reduces resource requirements and accelerates training.

\begin{wrapfigure}{r}{0.5\textwidth}
    \centering
    \includegraphics[width=\linewidth]{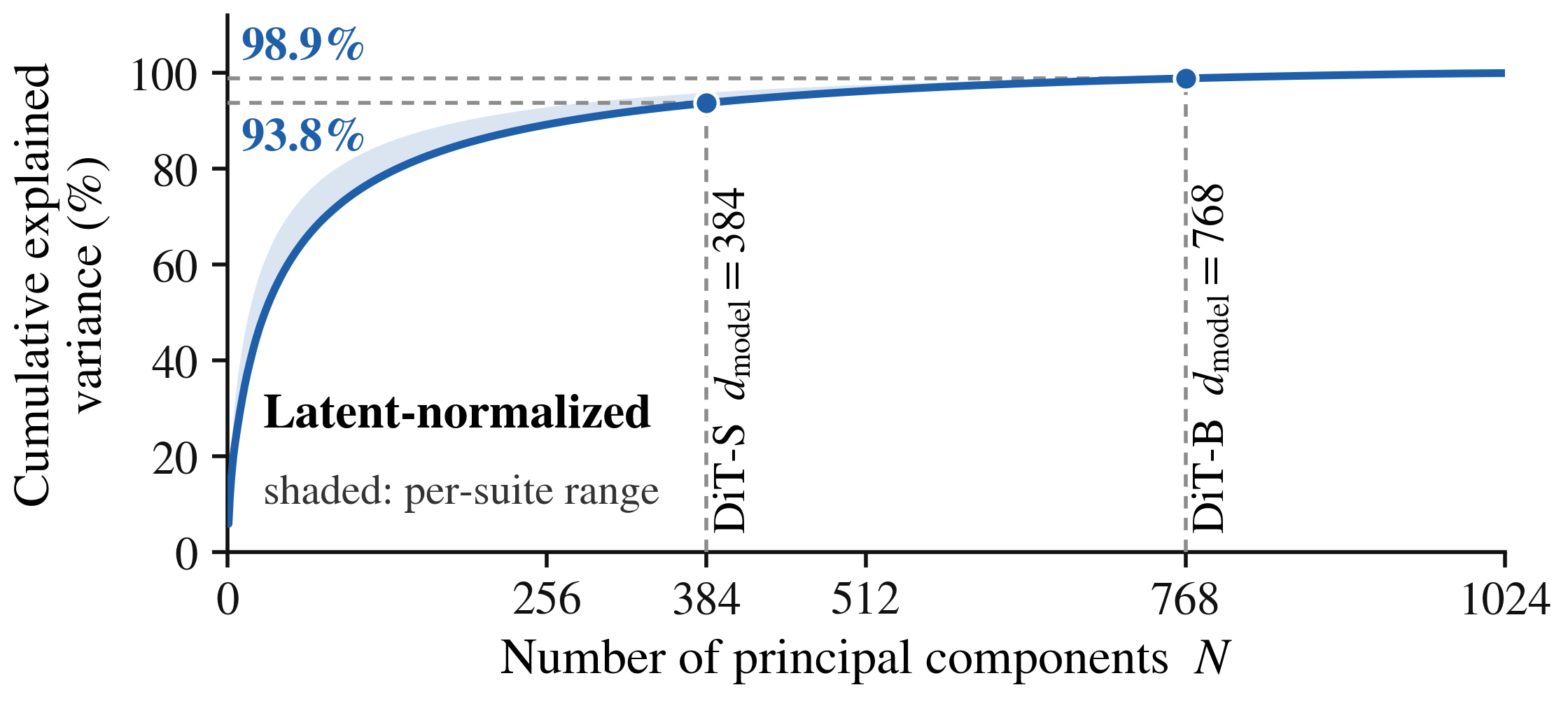}
    \vspace{-20pt}
    \caption{Cumulative explained variance of V-JEPA~2.1 latents. The shaded band is the range across benchmark suites. The dashed line marks the actualizer width $d_{\text{model}}=384$, at which $93.8\%$ of the variance is retained.}
    \label{fig:vjepa_energy}
    \vspace{-10pt}
\end{wrapfigure}

\subsection{Why a Small Actualizer can Work}
\label{sec:why_small}


Before turning to the architecture, we give a representational justification that a narrow actualizer is enough to estimate the distribution of each component in Eq.~\eqref{eq:mixture}. For a small actualizer with network width $d_{\text{model}}, d_{\text{model}}< D$, it can estimate a geometrical component with a rank of at most $d_{\text{model}}$. As a result, the representational ability of the actualizer is bound by the leading $n$ principal components in the representation space of $z_t$. Fig.~\ref{fig:vjepa_energy} evaluates this bound on the latents of Sec.~\ref{sec:representation} extracted from demonstrations in LIBERO~\cite{liu2023libero}. It shows that $d_{\text{model}}=384$ (width of DiT-S~\citep{peebles2023scalable}) components retain $93.8\%$ of the variance, so the width itself forces a loss of only $6.2\%$, and it is stable across benchmark suites (shaded band).

As a result, a narrow actualizer of DiT-S scale can nearly losslessly model the task-conditioned components of $\mathcal F(z_t)$ and read out actions. However, the na\"{i}ve \textit{v-prediction} target contains isotropic Gaussian noise of full rank $D$, which a narrow actualizer cannot capture. We therefore adopt the clean-sample parameterization (\aka \textit{x-prediction})~\citep{li2026back,singh2026improved}, in which the network predicts $\mathbf{z}^+_{t:t+H}$ itself, a target with the same low-rank structure as its input (Appendix~\ref{app:architecture_ablation}). 

\subsection{Lightweight Actualizer Model}
\label{sec:actualizer}

We implement $q_\theta$ as a joint flow-matching model over $(\mathbf z^+_{t:t+H},\mathbf a_{t:t+H})$, realized by two lightweight Diffusion Transformers (DiTs)~\cite{peebles2023scalable} organized as a Mixture-of-Transformers (MoT) architecture~\citep{liang2024mixture}. A \emph{latent expert} predicts the task-conditioned future in latent space along with an \emph{action expert} generates the action chunk that realizes the future. The two experts thus implement the determination and realization factors of Eq.~\eqref{eq:factors} separately. The experts have separate weights and interact through shared self-attention layers with block masks (Fig.~\ref{fig:attention_mask}). Observation tokens $z_t$ alongside the noisy future latent tokens enter the network through the latent expert, while the noisy action tokens are fed into the action expert. Both the future latent and actions attend to the intermediate features $h_t$ of clean latent. The task instruction is encoded by a frozen UMT5 text encoder~\citep{chung2023unimax}, which together with the encoded proprioceptive state $s_t$, conditions both experts via cross-attention layers. 

As explained at the end of Sec.~\ref{sec:why_small}, each expert predicts clean tokens as the \textit{x-prediction} target. Following the clean-sample parameterization of \citet{li2026back}, the latent expert and action expert predict $\hat{\mathbf{z}}^+_{t:t+H}$ and $\hat{\mathbf{a}}_{t:t+H}$ separately, but the training loss is computed in the velocity space: 
\begin{equation}
  \mathcal L=\mathcal L_a+\lambda_z\mathcal L_z,\qquad
  \mathcal L_a=\mathbb E\big\|(\mathbf{{a}}-\hat{\mathbf{a}})/(1-\tau_a)\big\|^2,\quad L_z=\mathbb E\big\|(\mathbf{{z}}^+-\hat{\mathbf{z}}^+)/(1-\tau_z)\big\|^2
\end{equation}
where the noisy tokens $\tilde{\mathbf{a}}_\tau=(1-\tau_a)\epsilon_a+\tau_a\mathbf{a}$ and $\tilde{\mathbf{z}}^+_\tau=(1-\tau_z)\epsilon_z+\tau_z\mathbf{z}^+$ with gaussian noise $\epsilon_a,\epsilon_z\sim\mathcal N(0,I)$ and noise level $\tau_a,\tau_z\in(0,1)$. $\lambda_z$ is a hyper-parameter to balance the two losses.

At each control step, the encoder is passed once to obtain $z_t$, and only the actualizer is iterated over $K$ denoising steps to produce the action chunk. With denoising confined to the small head, the policy runs at over 25\,Hz on a single NVIDIA  A6000 GPU.

\section{Experiments}
\label{sec:experiments}

We conduct extensive experiments to evaluate the performance of \model{} in both simulation and real environments. Sec.~\ref{sec:exp_setup} describes the benchmarks, baselines, and protocol. Sec.~\ref{sec:exp_sim} provides the simulation results including three benchmarks: LIBERO, LIBERO-Plus, and RoboTwin~2.0. Sec.~\ref{sec:exp_real} deploys \model{} on two physical platforms, a parallel-gripper arm and a bimanual system with dexterous hands. Sec.~\ref{sec:exp_cost} quantifies the training cost and inference latency. Finally, Sec.~\ref{sec:exp_ablations} gives a deep dive into how \model{} works and provides ablation studies.

\subsection{Experimental Setup}
\label{sec:exp_setup}

\noindent\textbf{Model Details.} Unless otherwise stated, we adopt a pretrained V-JEPA~2.1~\citep{mur2026v} (ViT-L/16, 300M) as the video world model encoder, while the actualizer is implemented with a DiT-S model (60M)~\citep{peebles2023scalable}. The actualizer is randomly initialized for each benchmark without any additional pretraining. More details are in the appendix.

\noindent\textbf{Benchmarks.} We evaluate \model{} on three simulation suites and two real-world platforms. The simulation benchmarks include: (\romannumeral 1) \textbf{LIBERO}~\citep{liu2023libero}: four task suites (Spatial, Object, Goal, Long) for robot manipulation; (\romannumeral 2) \textbf{LIBERO-Plus}~\citep{fei2025liberoplus}: a perturbed variant of LIBERO to evaluate the generalization ability in out-of-distribution situations; (\romannumeral 3) RoboTwin~2.0~\citep{chen2025robotwin}: a large-scale bimanual manipulation benchmark. We also conduct experiments on two different real-world robot hardware: (\romannumeral 1) \textbf{FANUC CRX-10iA arm}: we conduct two single-arm manipulation tasks; (\romannumeral 2) \textbf{Unitree G1}: we apply three bimanual dexterous manipulation tasks following prior work \citep{chen2026clift}.

\noindent\textbf{Baselines.} On simulation benchmarks, we compare against multiple previous works in the field of vision-language-action (VLA) model and world-action model (WAM). For real-world robots, we implement two representative robot policy baselines: $\pi_{0.5}$~\citep{zhou2025vision} as the VLA baseline, and Fast-WAM~\citep{yuan2026fast} as the WAM baseline.


\begin{table}[t]
\centering
\scriptsize
\setlength{\tabcolsep}{2.3pt}
\renewcommand{\arraystretch}{1.05}
\caption{
Success rates (\%) on LIBERO benchmark.
``Robot Pretrain'' denotes additional embodied data
pretraining before downstream task training.}
\label{tab:libero_main}
\vspace{-10pt}
\begin{tabular*}{\linewidth}{
@{\extracolsep{\fill}}
l|c|c|cccc|c|c
@{}
}
\toprule
Method &
\makecell{Robot\\Pretrain} &
\makecell{Trainable\\Parameters} &
Spatial &
Object &
Goal &
Long &
Overall &
Latency (ms)\\
\midrule

\grayrow{OpenVLA~\citep{kim2024openvla}
 & \Checkmark & 7.2B
& 84.7 & 88.4 & 79.2 & 53.7 & 76.5 & 145.5}\\

\grayrow{OpenVLA-OFT~\citep{kim2025fine}
 & \Checkmark & 7.2B
& 95.2 & 94.2 & 95.2 & 93.2 & 94.5 & 129.0} \\

\grayrow{NORA~\citep{hung2025nora}
&  \Checkmark & 3.8B 
& 92.2 & 95.4 & 89.4 & 74.6 & 87.9 & 190.4
}\\

\grayrow{$\pi_0$~\citep{black2024pi0}
& \Checkmark & 3.3B 
& 96.8 & 98.8 & 95.8 & 85.2 & 94.1 & 120.4}\\

\grayrow{$\pi_0$-Fast~\citep{pertsch2025fast}
& \Checkmark & 2.9B
& 96.4 & 96.8 & 88.6 & 60.2 & 85.5 & 68.5 
}\\

\grayrow{$\pi_{0.5}$~\citep{zhou2025vision}
& \Checkmark & 3.3B 
& 98.8 & 98.2 & 98.0 & 92.4 & 96.9 & 128.5}\\

\grayrow{UniVLA~\citep{bu2025univla}
& \Checkmark & 7.2B 
& 96.5 & 96.8 & 95.6 & 92.0 & 95.2 & 157.3}\\

\grayrow{RIPT-VLA~\cite{tan2025interactive}
& \Checkmark & 7.0B 
& 98.6 & 98.6 & 99.0 & 93.8 & 97.5 & 192.9} \\

\grayrow{Motus~\citep{bi2026motus}
& \Checkmark & 5.9B 
& 96.8 & 99.8 & 96.6
& 97.6 & 97.7 & 2230.0}\\

\grayrow{FOCA~\citep{nguyen2026foca}
& \Checkmark & 3.2B 
& 97.0 & 99.8 & 97.4
& 92.4 & 96.6 & 134.3 }\\

\midrule

WorldVLA~\citep{cen2025worldvla}
& \XSolidBrush & 7.2B 
& 87.6 & 96.2 & 83.4 & 60.0 & 81.8 & 397.5\\

Fast-WAM~\citep{yuan2026fast}
& \XSolidBrush & 6.0B 
& 98.2 & \best{100.0} & \best{97.0}
& 95.2 & 97.6 & 111.7\\

\midrule

\model{} (ours)
& \XSolidBrush & \best{0.06B} 
& \textbf{98.4}
& \best{100.0}
& 96.6
& \textbf{96.8}
& \best{98.0} & \best{39.9}\\

\bottomrule
\end{tabular*}
\vspace{-10pt}
\end{table}
\subsection{Generalization in Simulation}
\label{sec:exp_sim}

\noindent\textbf{LIBERO Benchmark.}
In Table~\ref{tab:libero_main}, we compare the performance of \model{} with baselines on LIBERO benchmark~\citep{liu2023libero}. We train a single \model{} model on the four standard evaluation suites (10 tasks per suite). With only 60M trainable parameters, \model{} achieves great performance with a 98.0\% average success rate. This performance outperforms multiple existing VLA or WAM baselines that have $100\times$ more trainable parameters.

\noindent\textbf{LIBERO-Plus Benchmark.}
LIBERO-Plus~\citep{fei2025liberoplus} is a more challenging benchmark built up on LIBERO tasks with notable visual and layout variations.
To evaluate \model{} on out-of-distribution situations, we only train \model{} on LIBERO training demonstrations without any additional data augmentations that fit the shifted distribution. Results in
Table~\ref{tab:libero_plus_main} reflect the strong generalization ability of \model{} despite its small trainable parameter count.
We suppose \model{} can exploit the strong generalization ability of the pretrained video world model, and only actualize the task-conditioned one with its limited network capacity.

\begin{table}[t]
\centering
\scriptsize
\setlength{\tabcolsep}{2.0pt}
\renewcommand{\arraystretch}{1.05}
\caption{
Success rates (\%) on LIBERO-Plus benchmark. Overall is the instance-weighted mean used by the LIBERO-Plus leaderboard.
$^*$: no open-source code to measure latency.}
\label{tab:libero_plus_main}
\vspace{-10pt}
\begin{tabular*}{\linewidth}{
@{\extracolsep{\fill}}
l|c|c|ccccccc|c|c
@{}
}
\toprule
Method &
\makecell{Robot\\Pretrain}&
\makecell{Trainable\\Parameters}&
Camera &
Robot &
Lang. &
Light &
Bg. &
Noise &
Layout &
Overall&
Latency (ms)\\
\midrule
\grayrow{OpenVLA~\citep{kim2024openvla}
& \Checkmark & 7.2B 
& 0.8 & 3.5 & 23.0 & 8.1
& 34.8 & 15.2 & 28.5 & 15.6 & 145.5} \\
\grayrow{OpenVLA-OFT~\citep{kim2025fine}
& \Checkmark & 7.2B 
& 56.4  & 31.9 & 79.5 &  88.7
& 93.3 & 75.8 &  74.2 & 69.6 & 129.0 } \\

\grayrow{NORA~\citep{hung2025nora}
&  \Checkmark & 3.8B 
& 2.2 & 37.0 & 65.1 & 45.7 
& 58.6 & 12.8 & 62.1 & 39.0 & 190.4}\\

\grayrow{$\pi_0$~\cite{black2024pi0}
& \Checkmark & 3.3B 
& 13.8 & 6.0 & 58.8 & 85.0
& 81.4 & 79.0 & 68.9 & 53.6 & 120.4}\\

\grayrow{$\pi_0$-Fast~\citep{pertsch2025fast}
& \Checkmark & 2.9B
& 65.1 & 21.6 & 61.0 & 73.2
& 73.2 & 74.4 & 68.8 & 61.6 &68.5} \\

\grayrow{UniVLA~\citep{bu2025univla}
& \Checkmark & 7.2B
& 1.8 & 46.2 & 69.6 & 69.0
& 81.0 & 21.2 & 31.9 & 42.9 & 157.3}\\

\grayrow{RIPT-VLA~\cite{tan2025interactive}
& \Checkmark & 7.0B
& 55.2 & 31.2 & 77.6 & 88.4 
& 91.6 & 73.5 & 74.2 & 68.4 & 192.9
}\\

\midrule

WorldVLA~\citep{cen2025worldvla}
& \XSolidBrush & 7.2B
& 0.1 & 27.9 & 41.6 & 43.7
& 17.1 & 10.9 & 38.0 & 25.0 & 397.5\\

Fast-WAM~\citep{yuan2026fast}
& \XSolidBrush & 6.0B
& 16.4 & 44.5 & 68.9 & 78.2
    & 53.7 & 37.7 & 60.7 & 50.0 & 111.7\\

DC-WAM~\citep{ji2026dc}
& \XSolidBrush & 6.0B
& 23.9 & 51.7
& \best{83.4}
& \best{91.7}
& \textbf{61.3}
& 54.2
& \best{69.8}
& 60.9 &$-^*$ \\

\midrule

\model{} (ours)
& \XSolidBrush & \best{0.06B} 
& \textbf{39.5} & \best{60.5} & 70.7 & 87.2
& 56.9 & \textbf{59.0} & 67.8 & \best{62.2} & \best{39.9}\\


\bottomrule
\end{tabular*}
\vspace{-10pt}
\end{table}

\begin{wraptable}{r}{0.45\textwidth}
\scriptsize
\vspace{-10pt}
\caption{Success rates (\%) on RoboTwin~2.0 (50 demos per task).}
\label{tab:robotwin_result}
\vspace{-10pt}
\resizebox{\linewidth}{!}{
\begin{tabular}{l|cc}
    \toprule
    Method & Trainable Params. & Success Rates (\%)\\
    \midrule
    $\pi_{0.5}$~\citep{zhou2025vision} & 3.3B & 31.4\\
    LingBot-VA~\citep{li2026causal} &5.3B & 17.2\\
    Fast-WAM~\citep{yuan2026fast} & 6.0B & 41.9\\
    \midrule
    \model{} (ours) & 60M & \textbf{58.8}\\
    \bottomrule
\end{tabular}
}
\vspace{-10pt}
\end{wraptable}

\noindent\textbf{RoboTwin~2.0 Benchmark (50 demos per task).}
We further evaluate \model{} under the RoboTwin~2.0 benchmark~\citep{chen2025robotwin}, in which tasks are much more challenging than LIBERO~\citep{liu2023libero}. Limited by computational resources, we randomly sample 50 demonstrations for each of the 50 tasks. This setting could also evaluate the generalization ability of \model{} with limited training data. As shown in Table~\ref{tab:robotwin_result}, \model{} greatly outperforms the baselines. Thanks to its small size, \model{} exhibits better data efficiency and avoids the overfitting issue.

\subsection{Real-World Deployment}
\label{sec:exp_real}

\begin{figure}[t]
  \centering
  \includegraphics[width=.8\linewidth]{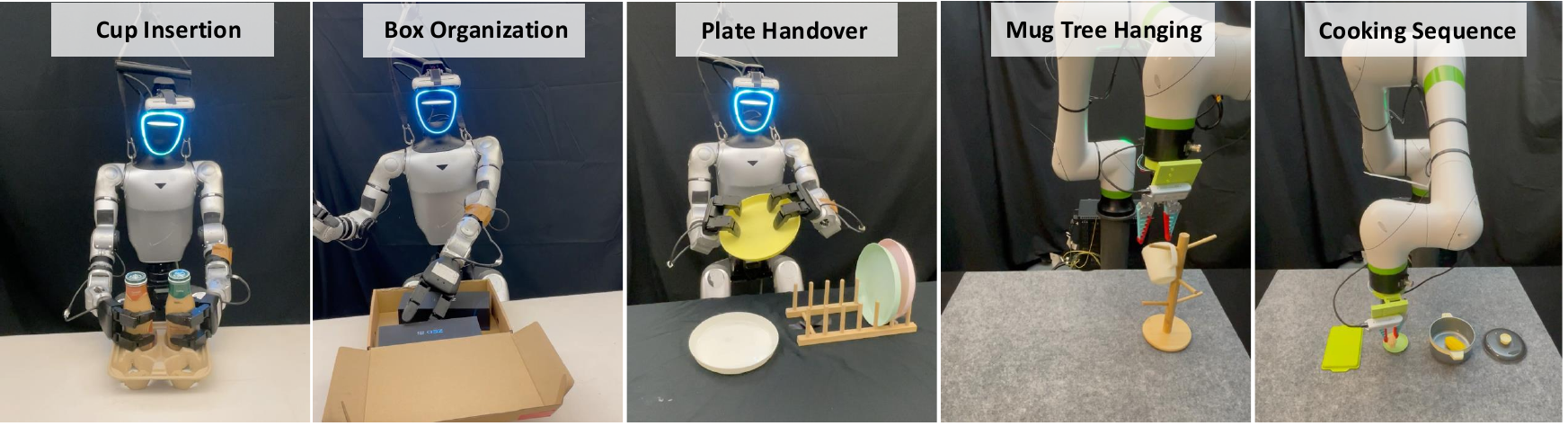}
  \vspace{-10pt}
  \caption{Real-world tasks on humanoid and robot arm platforms.}
  \label{fig:realworld}
  \vspace{-10pt}
\end{figure}

As shown in Fig.~\ref{fig:realworld}, we evaluated our method on five real-world tasks across two platforms: a G1 humanoid with a Unitree Dex3-1 hand for bimanual manipulation and a FANUC CRX-i10 arm with a gripper for single-arm manipulation. The G1 operated with its lower body stationary and used a head-mounted RealSense camera. The FANUC used two third-person cameras and one in-hand camera. We collected approximately 170 demonstrations per humanoid task and 100 per single-arm task, reflecting their differences in difficulty and action dimensionality (details in Appendix.~\ref{app:realworld}).
Fast-WAM and \model{} are trained from scratch, while $\pi_{0.5}$ is fine-tuned from its pretrained weight.
Table~\ref{tab:real_world} shows that our method consistently outperforms the corresponding baselines across all real-world tasks, demonstrating effective learning across single-arm and bimanual platforms. Videos are in the supplementary materials.

\subsection{Cost and Efficiency}
\label{sec:exp_cost}

\noindent\textbf{Single-GPU Model Training.}
We train \model{} on robot data on the top of cached V-JEPA latent features which relieves the burden of repetitive world model inference. We only optimize as few as 60M model parameters of the actualizer, which allows us to finish the training on a single NVIDIA PRO~6000 GPU with 32\,GB peak memory in only 15 hours on LIBERO Benchmark~\citep{liu2023libero}. In contrast, our WAM baseline, Fast-WAM~\citep{yuan2026fast}, has 6B parameters and requires an estimated 112 GPU-hours under its reported LIBERO training configuration.


\paragraph{Efficient Inference.}
In the deployment, the V-JEPA encoder only runs inference once, while the lightweight actualizer executes the repetitive denoising steps. Inspired by Fast-WAM~\citep{yuan2026fast}, the future latent prediction is discarded, so the actualizer only predicts the action and intermediate features of the clean frame can be cached in the denoising process. In Table~\ref{tab:libero_main} and Table~\ref{tab:libero_plus_main}, we measure the latency of all the methods on a single NVIDIA RTX A6000 GPU. \model{} yields a latency of 39\,ms, significantly faster than all the previous works. This enables \model{} to easily satisfy the requirements of real-time robot control. In Fig.~\ref{fig:latency_split}, most of \model{}'s latency is occupied by the one-time video encoder, and the denoising latency is very small.


\begin{table}[t]
\centering
\scriptsize
\begin{minipage}{0.7\linewidth}
\caption{Success rates (\%) in real-world experiments. 
}
\label{tab:real_world}
\vspace{-10pt}
\begin{subtable}{0.5\linewidth}
\centering
\caption{FANUC CRX-10iA (20 rollouts)}
\begin{tabular}{lcc}
\toprule
Method
& \shortstack{Mug}
& Cooking \\
\midrule
Fast-WAM
& 10 & 15 \\
\model{}
& \textbf{95}
& \textbf{60} \\
\bottomrule
\end{tabular}
\end{subtable}
\hfill
\begin{subtable}{0.5\linewidth}
\centering
\caption{Unitree G1 (30 rollouts)}
\begin{tabular}{lccc}
\toprule
Method
& \shortstack{Box}
& \shortstack{Cup}
& \shortstack{Plate} \\
\midrule
$\pi_{0.5}$
& 57 & 50 & 30 \\
\model{}
& \textbf{77}
& \textbf{83}
& \textbf{90} \\
\bottomrule
\end{tabular}
\end{subtable}
\end{minipage}
\hfill
\begin{minipage}{0.28\linewidth}
    \vspace{-5pt}
    \centering
    \caption{Ablation on latent-action joint prediction}
    \label{tab:joint_prediction}
    \begin{tabular}{cc}
    \toprule
        Latent Prediction & Success Rates (\%)\\
        \midrule
         \XSolidBrush &  17.9 \\
         \Checkmark &  \best{60.0} \\
    \bottomrule
    \end{tabular}
\end{minipage}
\vspace{-10pt}
\end{table}

\subsection{What Enables Lightweight Actualization?}
\label{sec:exp_ablations}
We study why a small actualizer can effectively read out a
frozen pretrained representation space, including qualitative and quantitative analysis. All the ablation studies are measured with success rates on LIBERO-Plus~\citep{fei2025liberoplus}. Due to some slightly different setups, the success rates are different with Table.~\ref{tab:libero_plus_main}, but the comparison fairness is guaranteed.

\subsubsection{How Lightweight Actualizer Helps?}
\label{sec:future_supervision}

\begin{figure}[t]
    \centering
    \includegraphics[width=0.9\linewidth]{
        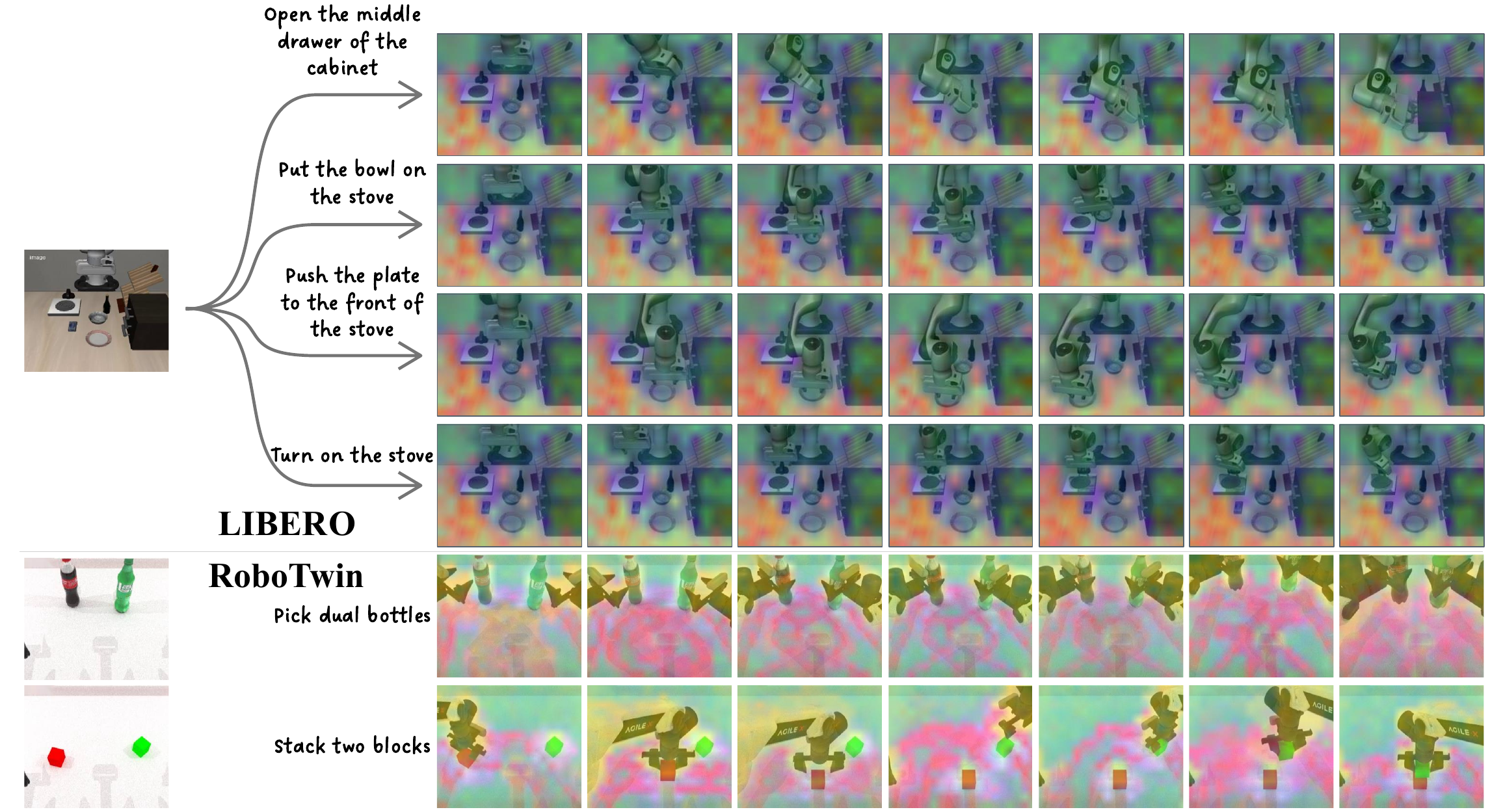
    }
    \caption{
    Predicted versus ground-truth future latent prediction at multiple time steps, visualized with PCA dimension reduction.
    Features predicted by latent expert tracks the teacher outputs greatly.
    }
    \label{fig:future_latent_alignment}
\end{figure}

Our proposed method actualizes the task-conditioned future and produces the actions to realize it. In Table~\ref{tab:joint_prediction}, removing the latent prediction target significantly reduces the success rate from 60\% to 17.9\%. This action-only prediction is equivalent to attaching a flow matching policy on top of the same frozen V-JEPA representation, which notably hurts the generalization ability. This reflects that the frozen V-JEPA representation itself is not strong enough for the robot policy but our joint prediction design fully stimulates the inherent future predictive ability from pretrained world models. Visualizations for future latent prediction are shown in Fig.~\ref{fig:future_latent_alignment}. The features are projected onto their first three PCA components
for RGB visualization. \model{} could predict both the task-conditioned future latent and diverse futures conditioned on different task instructions faithfully.

\begin{table}[t]
\centering
\scriptsize
\setlength{\tabcolsep}{4.2pt}
\renewcommand{\arraystretch}{1.08}
\caption{
Ablation study on latent representation. The ability for future prediction inside the latent space is more important than its representation ability.
}
\label{tab:representation}
\vspace{-10pt}
\begin{tabular}{lcccc}
\toprule
Representation &
Params. &
Input Range &
Latent Predictivity &
Success Rates (\%) \\

\midrule
WAN VAE & 127M & 4 frames & \XSolidBrush & 25.9 \\
\midrule
DINOv3 & 300M & 1 frame & \XSolidBrush & 39.7 \\
DINOv3 & 300M & 4 frames & \XSolidBrush & 29.8 \\
\midrule 

V-JEPA~$2.1_{image}$ & 300M & 1 frame & \XSolidBrush & 46.4 \\

V-JEPA~$2.1_{image}$ & 300M & 4 frames & \XSolidBrush & 44.8 \\

\midrule



V-JEPA~$2.1_{video}$ & 300M & 4 frames & \Checkmark & \best{60.0} \\

\bottomrule
\end{tabular}
\vspace{-10pt}
\end{table}

\subsubsection{How is \model{} Different from Latent World Action Models?}
\label{sec:representation_choice}

Several previous or concurrent works~\citep{wu2026pragmatic,yan2026flexpimultistreamworldactionmodel,lin2026jepa} have explored joint prediction of future latents and actions. However, \model{} comes from a fundamentally different motivations. Other works consider latent features as a superior alternative to RGB images due to their rich high-level semantics. As a result, they still require huge models with billions of parameters to learn the future predictions. In contrast, \model{} makes full use of the inherent future predictive ability inside the video world model representation space, so \model{} only determines rather than imagines task-conditioned futures. This largely simplifies the learning of our actualizer, and thus enables an extremely lightweight model. In Table~\ref{tab:representation}, we conduct an ablation study to deprive the future prediction ability by changing the latent representation space. Results show that other representation space (\eg DINO) brings bad performance despite its strong representative ability. Further, we adopt the same V-JEPA~2.1 encoder but apply to single image frame separately, which ensures the same representation space but strips the inherent future indication. This also brings inferior performance in Table~\ref{tab:representation}. We could conclude that the future prediction ability inside the latent is much more important than its representative ability. This also justifies our motivation to actualize the task-conditioned future from the diverse predictions rather than learn the future prediction that relieves the computation cost for training the actualizer.

\subsubsection{How should we allocate the computation?}
\label{sec:expert_parameterization}
As our model is composed of a pretrained video world model encoder and a trainable actualizer, it is useful to discuss how to allocate our computation. Based on the results in Table~\ref{tab:allocation}, we find it more beneficial to scale up the pretrained world model encoder rather than the trainable actualizer, although the world model is always frozen. This fact suggests that the future prediction ability of the frozen world model latent space is critical for the robot policy. We do not need a powerful actualizer, as it only needs to determine the task-conditioned future among those admitted by the world model rather than learning to predict futures by itself. Additional ablations of the architectural design are reported in Appendix~\ref{app:architecture_ablation}.

\begin{table}[t]
\centering
\begin{minipage}{0.5\linewidth}
\centering
\scriptsize
\captionof{table}{
Computation allocation between world model encoder and actualizer.
}
\label{tab:allocation}
\vspace{-10pt}
\begin{tabular}{lcc}
\toprule
World Model & Actualizer & Success Rates (\%) \\
\midrule
V-JEPA~2.1 (80M)  & DiT-S (60M)  & 50.7 \\
V-JEPA~2.1 (80M)  & DiT-B (245M)  & 53.2 \\
V-JEPA~2.1 (300M) & DiT-S (60M)  & 60.0 \\
V-JEPA~2.1 (300M) & DiT-B (245M) & \best{61.0} \\
\bottomrule
\end{tabular}
\end{minipage}
\hfill
\begin{minipage}{0.48\textwidth}
    \centering
    \includegraphics[width=\linewidth]{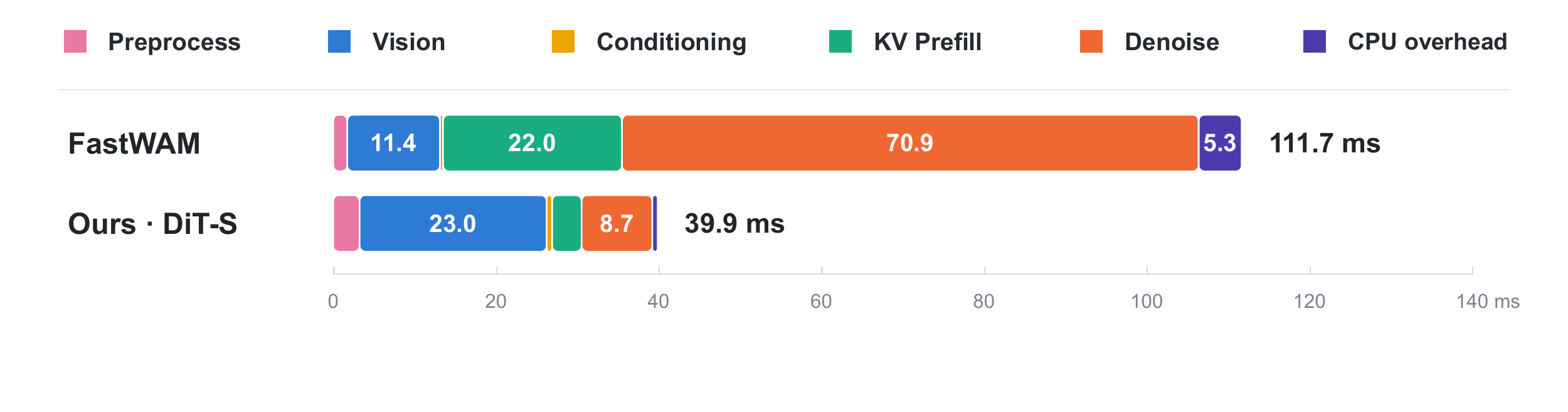}
    \vspace{-20pt}
    \captionof{figure}{Inference latency split of Fast-WAM and \model{}.}
    \label{fig:latency_split}
\end{minipage}
\vspace{-10pt}
\end{table}

\section{Conclusion}
\label{sec:conclusion}

We have presented \model{}, built on a division of labor: a frozen video world model encoder lays out the plausible futures of a scene in its latent space, and a small actualizer determines the task-conditioned one and realizes it as actions. Under this design, priors over future dynamics are inherited from the frozen representation space, and what must be learned from robot data reduces to a task-conditioned determination and a readout of actions, which yields a problem of far lower complexity than predicting scene evolution itself. We instantiate the actualizer as a compact DiT (60M trainable parameters) with an MoT architecture that jointly predicts latent features and actions conditioned on task instructions. Our experiments show that this small model outperforms VLA and WAM baselines with billions of trainable parameters on LIBERO, LIBERO-Plus, RoboTwin~2.0, and two real-world setups.

\noindent\textbf{Limitations.} Due to limited resources, we do not conduct experiments on the complete RoboTwin~2.0 benchmark~\citep{chen2025robotwin}. Besides, we only apply our method to V-JEPA~2.1 video world model encoder. The policy is bounded by the future predictive ability of world model latent space. We will list the combination of \model{} with other strong world model encoders and experiments on large-scale benchmarks in our future works.


\subsection*{AI Use Statement}
In this work, we used generative AI tools for implementing methods. We have not used generative AI tools for research ideation or the development of the core research ideas of this work, and generating synthetic datasets, formulating mathematical claims, and proving mathematical claims are not applicable to this work. Additionally, we used generative AI tools for creating and editing software code, creating research artifacts including tables and supplementary materials, drafting portions of the manuscript, and editing the paper for readability. We have reviewed all AI-assisted work. We verified the correctness of AI-assisted outputs and their consistency with our experiments and analyses. We take responsibility for the final content of this work, including text, claims or artifacts produced with the aid of generative AI.

\subsection*{Reproducibility Statement}
The main paper describes the observation construction, prediction targets, training objective, and hardware requirements. Appendix~\ref{app:implementation} summarizes the confirmed implementation settings. We will release the code when the paper is accepted.

\appendix


\bibliography{arxiv}

@article{mur2026v,
  title = {{V-JEPA 2.1}: Unlocking Dense Features in Video Self-Supervised Learning},
  author = {Mur-Labadia, Lorenzo and Muckley, Matthew and Bar, Amir and Assran, Mido and Sinha, Koustuv and Rabbat, Mike and LeCun, Yann and Ballas, Nicolas and Bardes, Adrien},
  journal = {arXiv preprint arXiv:2603.14482},
  year = {2026},
  url = {https://arxiv.org/abs/2603.14482}
}

@inproceedings{li2026back,
  title = {Back to Basics: Let Denoising Generative Models Denoise},
  author = {Li, Tianhong and He, Kaiming},
  booktitle = {Proceedings of the IEEE/CVF Conference on Computer Vision and Pattern Recognition},
  pages = {36115--36125},
  year = {2026}
}

@article{yuan2026fast,
  title = {{Fast-WAM}: Do World Action Models Need Test-time Future Imagination?},
  author = {Yuan, Tianyuan and Dong, Zibin and Liu, Yicheng and Zhao, Hang},
  journal = {arXiv preprint arXiv:2603.16666},
  year = {2026},
  url = {https://arxiv.org/abs/2603.16666}
}

@article{kim2024openvla,
  title = {{OpenVLA}: An Open-Source Vision-Language-Action Model},
  author = {Kim, Moo Jin and Pertsch, Karl and Karamcheti, Siddharth and Xiao, Ted and Balakrishna, Ashwin and Nair, Suraj and Rafailov, Rafael and Foster, Ethan and Lam, Grace and Sanketi, Pannag and Vuong, Quan and Kollar, Thomas and Burchfiel, Benjamin and Tedrake, Russ and Sadigh, Dorsa and Levine, Sergey and Liang, Percy and Finn, Chelsea},
  journal = {arXiv preprint arXiv:2406.09246},
  year = {2024},
  url = {https://arxiv.org/abs/2406.09246}
}

@article{black2024pi0,
  title = {{\ensuremath{\pi_0}}: A Vision-Language-Action Flow Model for General Robot Control},
  author = {Black, Kevin and Brown, Noah and Driess, Danny and Esmail, Adnan and Equi, Michael and Finn, Chelsea and Fusai, Niccolo and Groom, Lachy and Hausman, Karol and Ichter, Brian and Jakubczak, Szymon and Jones, Tim and Ke, Liyiming and Levine, Sergey and Li-Bell, Adrian and Mothukuri, Mohith and Nair, Suraj and Pertsch, Karl and Shi, Lucy Xiaoyang and Tanner, James and Vuong, Quan and Walling, Anna and Wang, Haohuan and Zhilinsky, Ury},
  journal = {arXiv preprint arXiv:2410.24164},
  year = {2024},
  url = {https://arxiv.org/abs/2410.24164}
}

@article{lin2026jepa,
  title = {{JEPA-WAM}: Learning Vision-Language-Action Policies with Joint-Embedding World Modeling},
  author = {Lin, Yihan and He, Jiawei and Bao, Shifeng and Zhao, Chen and Li, Yang and Wang, Xiaobo and Wang, Yan and Chi, Cheng and Zhang, Jing},
  journal = {arXiv preprint arXiv:2608.09381},
  year = {2026},
  url = {https://arxiv.org/abs/2608.09381}
}

@article{wang2026stwam,
  title = {{ST-WAM}: Semantic-Temporal World Action Model for Robust Manipulation under Visual Distribution Shifts},
  author = {Wang, Mingxin and Hu, Bin and Qian, Bin and Jiang, Kaitao and Wu, Haoning and Yan, Feng and Jing, Bowen and Hao, Ruiyang and Wang, Enyi and Niu, Kangning and Yang, Yandan and Xu, Mu and Wang, Yan and Liu, Houde and Li, Tianlun},
  journal = {arXiv preprint arXiv:2607.28993},
  year = {2026},
  url = {https://arxiv.org/abs/2607.28993}
}

@article{assran2025vjepa2,
  title = {{V-JEPA 2}: Self-Supervised Video Models Enable Understanding, Prediction and Planning},
  author = {Assran, Mahmoud and Bardes, Adrien and Fan, David and Garrido, Quentin and Howes, Russell and Komeili, Mojtaba and Muckley, Matthew and Rizvi, Ammar and Roberts, Claire and Sinha, Koustuv and Zholus, Artem and Arnaud, Sergio and Gejji, Abha and Martin, Ada and Hogan, Francois Robert and Dugas, Daniel and Bojanowski, Piotr and Khalidov, Vasil and Labatut, Patrick and Massa, Francisco and Szafraniec, Marc and Krishnakumar, Kapil and Li, Yong and Ma, Xiaodong and Chandar, Sarath and Meier, Franziska and LeCun, Yann and Rabbat, Michael and Ballas, Nicolas},
  journal = {arXiv preprint arXiv:2506.09985},
  year = {2025}
}

@article{pertsch2025fast,
  title = {{FAST}: Efficient Action Tokenization for Vision-Language-Action Models},
  author = {Pertsch, Karl and Stachowicz, Kyle and Ichter, Brian and Driess, Danny and Nair, Suraj and Vuong, Quan and Mees, Oier and Finn, Chelsea and Levine, Sergey},
  journal = {arXiv preprint arXiv:2501.09747},
  year = {2025},
  url = {https://arxiv.org/abs/2501.09747}
}

@article{bu2025univla,
  title = {{UniVLA}: Learning to Act Anywhere with Task-centric Latent Actions},
  author = {Bu, Qingwen and Yang, Yanting and Cai, Jisong and Gao, Shenyuan and Ren, Guanghui and Yao, Maoqing and Luo, Ping and Li, Hongyang},
  journal = {arXiv preprint arXiv:2505.06111},
  year = {2025},
  url = {https://arxiv.org/abs/2505.06111}
}

@article{cen2025worldvla,
  title = {{WorldVLA}: Towards Autoregressive Action World Model},
  author = {Cen, Jun and Yu, Chaohui and Yuan, Hangjie and others},
  journal = {arXiv preprint arXiv:2506.21539},
  year = {2025},
  url = {https://arxiv.org/abs/2506.21539}
}

@article{nair2022r3m,
  title = {{R3M}: A Universal Visual Representation for Robot Manipulation},
  author = {Nair, Suraj and Rajeswaran, Aravind and Kumar, Vikash and Finn, Chelsea and Gupta, Abhinav},
  journal = {arXiv preprint arXiv:2203.12601},
  year = {2022},
  url = {https://arxiv.org/abs/2203.12601}
}

@article{fei2025liberoplus,
  title = {{LIBERO-Plus}: In-depth Robustness Analysis of Vision-Language-Action Models},
  author = {Fei, Senyu and Wang, Siyin and Shi, Junhao and Dai, Zihao and Cai, Jikun and Qian, Pengfang and Ji, Li and He, Xinzhe and Zhang, Shiduo and Fei, Zhaoye and Fu, Jinlan and Gong, Jingjing and Qiu, Xipeng},
  journal = {arXiv preprint arXiv:2510.13626},
  year = {2025},
  url = {https://arxiv.org/abs/2510.13626}
}

@article{maes2026leworldmodel,
  title={Leworldmodel: Stable end-to-end joint-embedding predictive architecture from pixels},
  author={Maes, Lucas and Lidec, Quentin Le and Scieur, Damien and LeCun, Yann and Balestriero, Randall},
  journal={arXiv preprint arXiv:2603.19312},
  year={2026}
}

@article{zhou2024dino,
  title={Dino-wm: World models on pre-trained visual features enable zero-shot planning},
  author={Zhou, Gaoyue and Pan, Hengkai and LeCun, Yann and Pinto, Lerrel},
  journal={arXiv preprint arXiv:2411.04983},
  year={2024}
}

@article{ye2026world,
  title={World action models are zero-shot policies},
  author={Ye, Seonghyeon and Ge, Yunhao and Zheng, Kaiyuan and Gao, Shenyuan and Yu, Sihyun and Kurian, George and Indupuru, Suneel and Tan, You Liang and Zhu, Chuning and Xiang, Jiannan and others},
  journal={arXiv preprint arXiv:2602.15922},
  year={2026}
}

@article{bjorck2025gr00t,
  title={Gr00t n1: An open foundation model for generalist humanoid robots},
  author={Bjorck, Johan and Casta{\~n}eda, Fernando and Cherniadev, Nikita and Da, Xingye and Ding, Runyu and Fan, Linxi and Fang, Yu and Fox, Dieter and Hu, Fengyuan and Huang, Spencer and others},
  journal={arXiv preprint arXiv:2503.14734},
  year={2025}
}

@article{li2026causal,
  title={Causal world modeling for robot control},
  author={Li, Lin and Zhang, Qihang and Luo, Yiming and Yang, Shuai and Wang, Ruilin and Han, Fei and Yu, Mingrui and Gao, Zelin and Xue, Nan and Zhu, Xing and others},
  journal={arXiv preprint arXiv:2601.21998},
  year={2026}
}

@misc{yan2026flexpimultistreamworldactionmodel,
      title={Flex-$\pi$: A Multi-Stream World-Action Model with Compute Flexibility}, 
      author={Ge Yan and Jinghao Liu and Yuzhi Fan and Lei Cai and Minwen Liao and Jesse Zhang and Dieter Fox},
      year={2026},
      eprint={2608.10860},
      archivePrefix={arXiv},
      primaryClass={cs.RO},
      url={https://arxiv.org/abs/2608.10860}, 
}

@article{team2026advancing,
  title={Advancing open-source world models},
  author={Team, Robbyant and Gao, Zelin and Wang, Qiuyu and Zeng, Yanhong and Zhu, Jiapeng and Cheng, Ka Leong and Li, Yixuan and Wang, Hanlin and Xu, Yinghao and Ma, Shuailei and others},
  journal={arXiv preprint arXiv:2601.20540},
  year={2026}
}

@article{wan2025wan,
  title={Wan: Open and advanced large-scale video generative models},
  author={Wan, Team and Wang, Ang and Ai, Baole and Wen, Bin and Mao, Chaojie and Xie, Chen-Wei and Chen, Di and Yu, Feiwu and Zhao, Haiming and Yang, Jianxiao and others},
  journal={arXiv preprint arXiv:2503.20314},
  year={2025}
}

@article{liang2024mixture,
  title={Mixture-of-transformers: A sparse and scalable architecture for multi-modal foundation models},
  author={Liang, Weixin and Yu, Lili and Luo, Liang and Iyer, Srinivasan and Dong, Ning and Zhou, Chunting and Ghosh, Gargi and Lewis, Mike and Yih, Wen-tau and Zettlemoyer, Luke and others},
  journal={arXiv preprint arXiv:2411.04996},
  year={2024}
}

@article{liu2023libero,
  title={Libero: Benchmarking knowledge transfer for lifelong robot learning},
  author={Liu, Bo and Zhu, Yifeng and Gao, Chongkai and Feng, Yihao and Liu, Qiang and Zhu, Yuke and Stone, Peter},
  journal={Advances in Neural Information Processing Systems},
  volume={36},
  pages={44776--44791},
  year={2023}
}

@article{chen2025robotwin,
  title={Robotwin 2.0: A scalable data generator and benchmark with strong domain randomization for robust bimanual robotic manipulation},
  author={Chen, Tianxing and Chen, Zanxin and Chen, Baijun and Cai, Zijian and Liu, Yibin and Li, Zixuan and Liang, Qiwei and Lin, Xianliang and Ge, Yiheng and Gu, Zhenyu and others},
  journal={arXiv preprint arXiv:2506.18088},
  year={2025}
}

@article{chen2026clift,
  title={CLIFT: Turning Gemini Robotics On-Device into Humanoid Specialists via Non-Invasive Closed-Loop Iterative Fine-Tuning},
  author={Chen, Yuxin and Srikanth, Hari and Jew, Nathan and Wu, Menglin and Wang, Pengcheng and Ren, Junli and Tomizuka, Masayoshi and Xu, Peng and Xie, Jinyu and Tian, Thomas},
  journal={arXiv preprint arXiv:2607.29172},
  year={2026}
}

@misc{openai2024sora,
  author       = {OpenAI},
  title        = {Video generation models as world simulators},
  howpublished = {\url{https://openai.com/index/video-generation-models-as-world-simulators/}},
  year         = {2024},
  note         = {Accessed: 2026-09-22}
}

@article{agarwal2025cosmos,
  title={Cosmos world foundation model platform for physical ai},
  author={Agarwal, Niket and Ali, Arslan and Bala, Maciej and Balaji, Yogesh and Barker, Erik and Cai, Tiffany and Chattopadhyay, Prithvijit and Chen, Yongxin and Cui, Yin and Ding, Yifan and others},
  journal={arXiv preprint arXiv:2501.03575},
  year={2025}
}

@article{bardes2024revisiting,
  title={Revisiting feature prediction for learning visual representations from video},
  author={Bardes, Adrien and Garrido, Quentin and Ponce, Jean and Chen, Xinlei and Rabbat, Michael and LeCun, Yann and Assran, Mahmoud and Ballas, Nicolas},
  journal={arXiv preprint arXiv:2404.08471},
  year={2024}
}

@inproceedings{bar2025navigation,
  title={Navigation world models},
  author={Bar, Amir and Zhou, Gaoyue and Tran, Danny and Darrell, Trevor and LeCun, Yann},
  booktitle={2025 IEEE/CVF Conference on Computer Vision and Pattern Recognition (CVPR)},
  pages={15791--15801},
  year={2025},
  organization={IEEE}
}

@article{oquab2023dinov2,
  title={Dinov2: Learning robust visual features without supervision},
  author={Oquab, Maxime and Darcet, Timoth{\'e}e and Moutakanni, Th{\'e}o and Vo, Huy and Szafraniec, Marc and Khalidov, Vasil and Fernandez, Pierre and Haziza, Daniel and Massa, Francisco and El-Nouby, Alaaeldin and others},
  journal={arXiv preprint arXiv:2304.07193},
  year={2023}
}

@article{du2023learning,
  title={Learning universal policies via text-guided video generation},
  author={Du, Yilun and Yang, Sherry and Dai, Bo and Dai, Hanjun and Nachum, Ofir and Tenenbaum, Josh and Schuurmans, Dale and Abbeel, Pieter},
  journal={Advances in neural information processing systems},
  volume={36},
  pages={9156--9172},
  year={2023}
}

@article{cheang2024gr,
  title={Gr-2: A generative video-language-action model with web-scale knowledge for robot manipulation},
  author={Cheang, Chi-Lam and Chen, Guangzeng and Jing, Ya and Kong, Tao and Li, Hang and Li, Yifeng and Liu, Yuxiao and Wu, Hongtao and Xu, Jiafeng and Yang, Yichu and others},
  journal={arXiv preprint arXiv:2410.06158},
  year={2024}
}

@article{hu2024video,
  title={Video prediction policy: A generalist robot policy with predictive visual representations},
  author={Hu, Yucheng and Guo, Yanjiang and Wang, Pengchao and Chen, Xiaoyu and Wang, Yen-Jen and Zhang, Jianke and Sreenath, Koushil and Lu, Chaochao and Chen, Jianyu},
  journal={arXiv preprint arXiv:2412.14803},
  year={2024}
}

@inproceedings{sun2026vla,
  title={Vla-jepa: Enhancing vision-language-action model with latent world model},
  author={Sun, Jingwen and Zhang, Wenyao and Qi, Zekun and Ren, Shaojie and Liu, Zezhi and Zhu, Hanxin and Sun, Guangzhong and Jin, Xin and Chen, Zhibo},
  booktitle={European Conference on Computer Vision},
  pages={478--497},
  year={2026},
  organization={Springer}
}

@article{zheng2025flare,
  title={Flare: Robot learning with implicit world modeling},
  author={Zheng, Ruijie and Wang, Jing and Reed, Scott and Bjorck, Johan and Fang, Yu and Hu, Fengyuan and Jang, Joel and Kundalia, Kaushil and Lin, Zongyu and Magne, Loic and others},
  journal={arXiv preprint arXiv:2505.15659},
  year={2025}
}

@article{shang2024theia,
  title={Theia: Distilling diverse vision foundation models for robot learning},
  author={Shang, Jinghuan and Schmeckpeper, Karl and May, Brandon B and Minniti, Maria Vittoria and Kelestemur, Tarik and Watkins, David and Herlant, Laura},
  journal={arXiv preprint arXiv:2407.20179},
  year={2024}
}

@inproceedings{xia2025cage,
  title={Cage: Causal attention enables data-efficient generalizable robotic manipulation},
  author={Xia, Shangning and Fang, Hongjie and Lu, Cewu and Fang, Hao-Shu},
  booktitle={2025 IEEE International Conference on Robotics and Automation (ICRA)},
  pages={13242--13249},
  year={2025},
  organization={IEEE}
}

@article{xie2026multi,
  title={Multi-Camera View Scaling for Data-Efficient Robot Imitation Learning},
  author={Xie, Yichen and Wang, Yixiao and Zhao, Shuqi and Wu, Cheng-En and Tomizuka, Masayoshi and Xie, Jianwen and Fang, Hao-Shu},
  journal={arXiv preprint arXiv:2604.00557},
  year={2026}
}

@inproceedings{wang2026ver,
  title={Ver: Vision expert transformer for robot learning via foundation distillation and dynamic routing},
  author={Wang, Yixiao and Huo, Mingxiao and Liang, Zhixuan and Du, Yushi and Sun, Lingfeng and Lin, Haotian and Shang, Jinghuan and Peng, Chensheng and Bansal, Mohit and Ding, Mingyu and others},
  booktitle={International Conference on Learning Representations},
  volume={2026},
  pages={155122--155148},
  year={2026}
}

@inproceedings{parisi2022unsurprising,
  title={The unsurprising effectiveness of pre-trained vision models for control},
  author={Parisi, Simone and Rajeswaran, Aravind and Purushwalkam, Senthil and Gupta, Abhinav},
  booktitle={international conference on machine learning},
  pages={17359--17371},
  year={2022},
  organization={PMLR}
}

@inproceedings{goodwin2022zero,
  title={Zero-shot category-level object pose estimation},
  author={Goodwin, Walter and Vaze, Sagar and Havoutis, Ioannis and Posner, Ingmar},
  booktitle={European Conference on Computer Vision},
  pages={516--532},
  year={2022},
  organization={Springer}
}

@article{simeoni2025dinov3,
  title={Dinov3},
  author={Sim{\'e}oni, Oriane and Vo, Huy V and Seitzer, Maximilian and Baldassarre, Federico and Oquab, Maxime and Jose, Cijo and Khalidov, Vasil and Szafraniec, Marc and Yi, Seungeun and Ramamonjisoa, Micha{\"e}l and others},
  journal={arXiv preprint arXiv:2508.10104},
  year={2025}
}

@inproceedings{peebles2023scalable,
  title={Scalable diffusion models with transformers},
  author={Peebles, William and Xie, Saining},
  booktitle={2023 IEEE/CVF International Conference on Computer Vision (ICCV)},
  pages={4172--4182},
  year={2023},
  organization={IEEE}
}

@article{singh2026improved,
  title={Improved baselines with representation autoencoders},
  author={Singh, Jaskirat and Zheng, Boyang and Wu, Zongze and Zhang, Richard and Shechtman, Eli and Xie, Saining},
  journal={arXiv preprint arXiv:2605.18324},
  year={2026}
}

@inproceedings{chung2023unimax,
title={UniMax: Fairer and More Effective Language Sampling for Large-Scale Multilingual Pretraining},
author={Hyung Won Chung and Xavier Garcia and Adam Roberts and Yi Tay and Orhan Firat and Sharan Narang and Noah Constant},
booktitle={The Eleventh International Conference on Learning Representations },
year={2023},
url={https://openreview.net/forum?id=kXwdL1cWOAi}
}

@article{zhou2025vision,
  title={Vision-language-action model with open-world embodied reasoning from pretrained knowledge},
  author={Zhou, Zhongyi and Zhu, Yichen and Wen, Junjie and Shen, Chaomin and Xu, Yi},
  journal={arXiv e-prints},
  pages={arXiv--2505},
  year={2025}
}

@inproceedings{bi2026motus,
  title={Motus: A unified latent action world model},
  author={Bi, Hongzhe and Tan, Hengkai and Xie, Shenghao and Wang, Zeyuan and Huang, Shuhe and Liu, Haitian and Zhao, Ruowen and Feng, Yao and Xiang, Chendong and Rong, Yinze and others},
  booktitle={Proceedings of the IEEE/CVF Conference on Computer Vision and Pattern Recognition},
  pages={35101--35113},
  year={2026}
}

@article{ji2026dc,
  title={DC-WAM: Dynamic-Centric Visual Supervision and Reasoning for World-Action Models},
  author={Ji, Haoyuan and Fan, Lingxiang and Su, Shang and Lu, Yinqiao and Shi, Mengkai and Gao, Jun and Feng, Shuo},
  journal={arXiv preprint arXiv:2607.25918},
  year={2026}
}

@article{wu2026pragmatic,
  title={A pragmatic vla foundation model},
  author={Wu, Wei and Lu, Fan and Wang, Yunnan and Yang, Shuai and Liu, Shi and Wang, Fangjing and Zhu, Qian and Sun, He and Wang, Yong and Ma, Shuailei and others},
  journal={arXiv preprint arXiv:2601.18692},
  year={2026}
}

@inproceedings{zheng2026diffusion,
  title={Diffusion transformers with representation autoencoders},
  author={Zheng, Boyang and Ma, Nanye and Tong, Shengbang and Xie, Saining},
  booktitle={International Conference on Learning Representations},
  volume={2026},
  pages={35791--35820},
  year={2026}
}

@article{kim2025fine,
  title={Fine-tuning vision-language-action models: Optimizing speed and success},
  author={Kim, Moo Jin and Finn, Chelsea and Liang, Percy},
  journal={arXiv preprint arXiv:2502.19645},
  year={2025}
}

@article{hung2025nora,
  title={Nora: A small open-sourced generalist vision language action model for embodied tasks},
  author={Hung, Chia-Yu and Sun, Qi and Hong, Pengfei and Zadeh, Amir and Li, Chuan and Tan, U and Majumder, Navonil and Poria, Soujanya and others},
  journal={arXiv preprint arXiv:2504.19854},
  year={2025}
}

@article{tan2025interactive,
  title={Interactive post-training for vision-language-action models},
  author={Tan, Shuhan and Dou, Kairan and Zhao, Yue and Kr{\"a}henb{\"u}hl, Philipp},
  journal={arXiv preprint arXiv:2505.17016},
  year={2025}
}

@article{nguyen2026foca,
  title={FOCA: Future-Oriented Conditioning for Data-Efficient Vision-Language-Action Adaptation},
  author={Nguyen, Duc Minh and Diep, Nghiem Tuong and Nguyen, Binh Gia and Ho, Trong-Bao and Le, Doanh and Nguyen, Tan Q and Ha, Thien-Loc and Tran, Nhiem and Thach, Bao and Tran, Nhat X and others},
  journal={arXiv preprint arXiv:2606.20867},
  year={2026}
}
\bibliographystyle{iclr2027_conference}
\clearpage
\appendix

\section{Additional Architecture Ablations}
  \label{app:architecture_ablation}
  
  \noindent\textbf{Mixture-of-Transformers Architecture.}
  We ablate the MoT architecture. The single-DiT variant shares weights across the latent and action experts, while keeping the other design choices
  unchanged. Table~\ref{tab:ablation_mot} shows that this substantially reduces performance: the LIBERO-Plus success rate drops from 60.0\% to 51.8\%. We
  hypothesize that the MoT architecture better separates the determination and realization processes in Eq.~\refeq{eq:factors}.

  \noindent\textbf{Prediction Target.}
  Table~\ref{tab:ablation_xpred} shows that the \textit{x-prediction} target is important for strong performance. This is consistent with our analysis in
  Sec.~\ref{sec:why_small}: \textit{x-prediction} allows the latent space to be represented with a narrow network.

  \noindent\textbf{Wide Head.}
  Given the high-dimensional latent space of V-JEPA~2.1, one possible approach is to use a wide head, as in
  RAE~\citep{zheng2026diffusion,singh2026improved}. However, this design does not improve \model{} (Table~\ref{tab:ablation_widehead}). As discussed in
  Sec.~\ref{sec:why_small}, the small actualizer is sufficient for actualization, and a wide head may make the training process unstable.

\begin{table}[h]
\centering
\scriptsize

\caption{Ablation study on model architecture design.}
\label{tab:design_ablation}

\vspace{-10pt}

\begin{subtable}[t]{0.35\linewidth}
\centering
\caption{Mixture of Transformers}
\label{tab:ablation_mot}
\vspace{-5pt}
\begin{tabular}{lc}
\toprule
Method & Success Rate (\%)\\
\midrule
Single DiT
& 51.8\\
MoT
& \best{60.0}\\
\bottomrule
\end{tabular}
\end{subtable}
\hfill
\begin{subtable}[t]{0.35\linewidth}
\centering
\caption{Prediction target}
\label{tab:ablation_xpred}
\vspace{-5pt}
\begin{tabular}{lc}
\toprule
Target  & Success Rate (\%)\\
\midrule
$v$-prediction
& 52.7\\
$x$-prediction
& \best{60.0}\\
\bottomrule
\end{tabular}
\end{subtable}
\hfill
\begin{subtable}[t]{0.27\linewidth}
\centering
\caption{Wide head}
\label{tab:ablation_widehead}
\vspace{-5pt}
\begin{tabular}{lc}
\toprule
Variant & Success Rate (\%)\\
\midrule
\textit{w/} wide head
& 48.5\\
\textit{w/o} wide head
& \best{60.0}\\
\bottomrule
\end{tabular}
\end{subtable}
\end{table}

\section{Implementation and Reproducibility Details}
\label{app:implementation}

\subsection{Model Architecture Details}
\label{app:architecture_details}
\paragraph{Frozen world model encoder.} All experiments use a frozen V-JEPA~2.1 ViT-L/16-384
encoder ($\sim$300M parameters, distilled from ViT-G), kept in \texttt{float32} and in
\texttt{eval} mode at all times; it is excluded from the optimizer and from the released
checkpoints.  Inputs are mapped from the $[-1,1]$
convention used throughout our pipeline to ImageNet normalization inside the backbone
wrapper. A frame is encoded by the \emph{video} branch (tubelet $2$) rather than the
single-image branch; the clip construction is described in
Sec.~\ref{app:observation_target}.

\begin{figure}[t]
  \centering
  \includegraphics[width=0.65\linewidth]{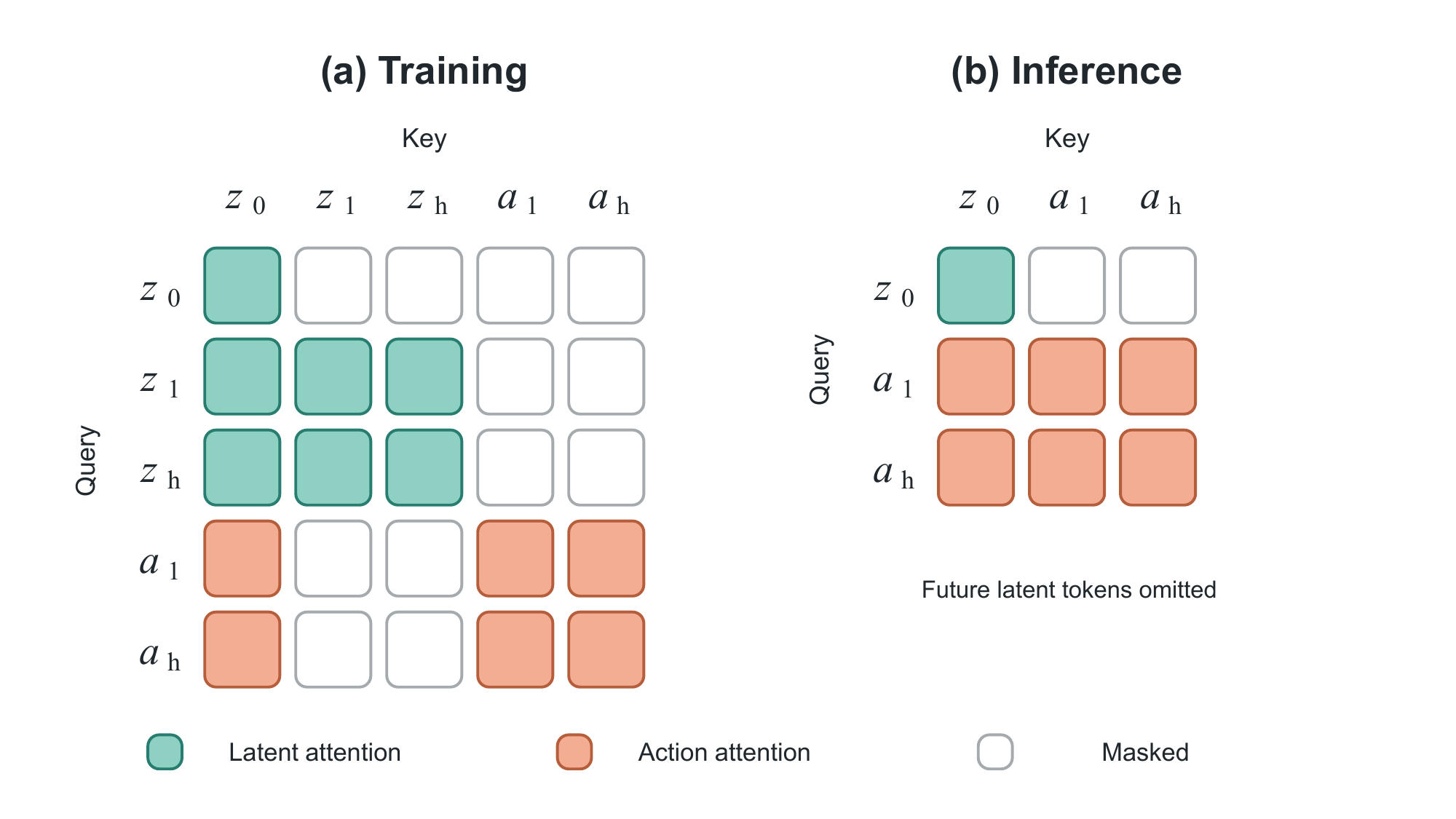}
  \caption{Block attention mask for joint self-attention in the DiTs.
  Current latent tokens attend only to current latent tokens.
  Future latent tokens attend to current and future latent tokens.
  Action tokens attend to current latent and action tokens.
  The action stream can therefore run without future latent prediction at inference.}
  \label{fig:attention_mask}
\end{figure}

\paragraph{Denoiser.} The denoiser is a two-branch mixture-of-transformers: an
\emph{image branch} and an \emph{action branch}, each with $12$ pre-norm blocks. At every
layer the two branches compute their own $q,k,v$ and are then mixed by a
\emph{joint self-attention with block mask} (Fig.~\ref{fig:attention_mask}) over the concatenated token stream, after which each
branch independently applies cross-attention to its own conditioning tokens and its own
FFN. Blocks use AdaLN-single conditioning (Wan style): one zero-initialized
$\mathrm{SiLU}\!\to\!\mathrm{Linear}(d,6d)$ head per branch is shared by all $12$ layers,
and each layer adds a learnable constant $\mathbf{m}\in\mathbb{R}^{6\times d}$, yielding
shift/scale/gate for attention and FFN. Attention uses RMSNorm on $q$ and $k$
($\epsilon=10^{-6}$), FFNs are $\mathrm{Linear}\!\to\!\mathrm{GELU}\!\to\!\mathrm{Linear}$
with ratio $4$, the pre-cross-attention LayerNorm is affine while the AdaLN-modulated
LayerNorms are affine-free, dropout is $0$, and no long skip connections are used.
Two sizes follow the official DiT configurations (depth $12$, head dim $64$):

\begin{center}
\begin{tabular}{lccccr}
\toprule
 & $d_{\text{model}}$ & heads & FFN & joint expert & trainable params \\
\midrule
DiT-S & 384 & 6  & 1536 & 56.80M  & 63,604,487 \\
DiT-B & 768 & 12 & 3072 & 226.84M & 245,762,567 \\
\bottomrule
\end{tabular}
\end{center}

\noindent The remaining parameters are the per-branch context encoders
($2\times1.87$M / $2\times4.33$M), timestep embedders ($2\times1.13$M / $2\times4.33$M),
the shared latent projection, and the two output heads.

\paragraph{Positional encoding.} Positions are encoded exclusively by RoPE (no additive
embeddings). With head dimension $64$ (32 complex pairs), image tokens use 3D RoPE split
$12/10/10$ pairs over (frame, height, width); the frame axis carries the \emph{actual}
environment step, $[0,4,8,12,16]$ for $z_0$ and the four targets. Action tokens use 1D
RoPE at positions $0,\dots,15$, so that action step $k$ and image frame $j$ live on the
same step axis. Rotations are applied to $q,k$ (never $v$) after qk-normalization, in
\texttt{float64}, then cast back.

\paragraph{Conditioning.} Language embeddings are precomputed once with the frozen
umT5-XXL text encoder of Wan2.2-TI2V-5B: $128$ tokens $\times$ $4096$ dims, using the
prompt template \emph{``A video recorded from a robot's point of view executing the
following instruction: \{task\}''}. Padded positions are zeroed and masked out of
cross-attention. Proprioception ($8$-D) is embedded by
$\mathrm{Linear}\!\to\!\mathrm{GELU}\!\to\!\mathrm{Linear}$ into one token appended to the
text tokens, giving $129$ context tokens. The two branches own \emph{separate} text and
proprio encoders (no weight sharing); a single conditioning stream is shared across all
$12$ layers of a branch.

\paragraph{Output heads.} The image head is
$\mathrm{LN}_{\text{no-affine}}\!\to\!\text{AdaLN modulate}\!\to\!\mathrm{Linear}(d,1024)$,
where shift/scale are a learnable $[2,d]$ constant plus the timestep embedding; the action
head is a single $\mathrm{Linear}(d,7)$. Both heads are zero-initialized, so the network
predicts $\hat{x}=0$ at initialization.

\paragraph{Latent normalization.} Before entering the denoiser, every V-JEPA latent is
normalized per channel, $\tilde{z}=(z-\mu)/\sigma$ with $\mu,\sigma\in\mathbb{R}^{1024}$
computed in \texttt{float64} over the entire cache ($1.11\times10^{11}$ stored fp16
values; global std $1.84$, per-channel $|\mu|$ up to $24$). The statistics are registered
as buffers and travel with the checkpoint, so training and evaluation provably share one
set of constants. The same transform is applied to $z_0$ at inference.
\subsection{Observation and Target Construction}
\label{app:observation_target}

\paragraph{Data.} We use the four LIBERO suites (Spatial, Object, Goal, Long) in their
no-noops LeRobot conversion at $20$\,Hz: $1{,}712$ demonstrations and $277{,}713$ frames
(mean episode length $122.6$/$147.3$/$122.2$/$268.8$ respectively). One training sample is
anchored at every frame (stride $1$), so the training set contains $277{,}713$ windows.

\paragraph{Image pipeline.} Raw $512\times512$ frames are decoded, converted to tensors and
resized per camera to $224\times224$, concatenated horizontally, then resized/center-cropped
to $224\times448$ and normalized to $[-1,1]$ --- the identical pipeline is used for cache
generation, online training and closed-loop evaluation.

\paragraph{Context latent $z_0$.} $z_0$ is not a single-frame embedding: the clip
$\{t-6,t-4,t-2,t\}$ (\texttt{clip\_len}$=4$, \texttt{frame\_interval}$=2$) is passed through
the V-JEPA \emph{video} branch (tubelet $2$), and the last temporal group is kept, giving
$392\times1024$ tokens that carry short-horizon motion while remaining index-aligned with
the action/proprio row of frame $t$. Indices before the episode start are clamped to the
first valid frame (i.e. the earliest frame is repeated).

\paragraph{Prediction targets.} The window spans $17$ frames; with
\texttt{action\_video\_freq\_ratio}$=4$ the image branch supervises the latents at
$t\!+\!4,t\!+\!8,t\!+\!12,t\!+\!16$ (\texttt{future\_size}$=4$), each encoded with the same
causal-clip rule as $z_0$. The action branch supervises the next $16$ normalized actions
(\texttt{chunk\_size}$=16$), i.e. the action chunk and the image targets cover the same
$16$-step horizon.

\paragraph{Action and state spaces.} Actions are $7$-D: a $6$-D delta end-effector pose
(OSC) plus an absolute gripper command. Proprioception is $8$-D: EE pose ($6$) plus the two
gripper joints, taken at the anchor frame only. Both are normalized per dimension to
$[-1,1]$ with min/max statistics computed over the training split (the normalizer clamps to
$\pm5$; it is exactly invertible for evaluation). For padded steps the six delta dimensions
are zeroed before normalization.

\subsection{Training Details}
\label{app:training_details}
\paragraph{Objective.}
Following Eq.~(6), both experts are trained with a shifted rectified-flow schedule
with $1000$ discrete noise levels. The noise level $\tau\in(0,1)$ ($\tau=1$ clean,
$\tau=0$ pure noise) is obtained from $u\sim\mathcal{U}(0,1)$ via
$1-\tau=\varphi(u)=su/(1+(s-1)u)$ with $s=5$ for both branches. Noisy tokens are formed as
$\tilde{\mathbf{a}}_\tau=(1-\tau_a)\boldsymbol{\epsilon}_a+\tau_a\mathbf{a}$ and
$\tilde{\mathbf{z}}^{+}_\tau=(1-\tau_z)\boldsymbol{\epsilon}_z+\tau_z\mathbf{z}^{+}$,
with $\boldsymbol{\epsilon}_a,\boldsymbol{\epsilon}_z\sim\mathcal{N}(0,I)$.
The action expert and the latent expert predict the clean tokens $\hat{\mathbf{a}}$ and
$\hat{\mathbf{z}}^{+}$, respectively, and the losses are computed in velocity space with a
clipped denominator,
\begin{equation}
\mathcal{L}_a=\mathbb{E}\,\Big\|\frac{\mathbf{a}-\hat{\mathbf{a}}}{\max(1-\tau_a,\,0.05)}\Big\|^{2},
\qquad
\mathcal{L}_z=\mathbb{E}\,\Big\|\frac{\mathbf{z}^{+}-\hat{\mathbf{z}}^{+}}{\max(1-\tau_z,\,0.05)}\Big\|^{2},
\end{equation}
i.e., the predicted velocity
$\hat{\mathbf{v}}=(\tilde{\mathbf{a}}_\tau-\hat{\mathbf{a}})/\max(1-\tau_a,0.05)$ is regressed
onto the target $\mathbf{v}=(\tilde{\mathbf{a}}_\tau-\mathbf{a})/\max(1-\tau_a,0.05)$, and
likewise for $\mathbf{z}^{+}$. Since prediction and target share the same clamped denominator,
this is exactly a $1/(1-\tau)^{2}$-weighted $x$-prediction loss with a floor at $1-\tau=0.05$.
The noise levels $\tau_a$ and $\tau_z$ are drawn \emph{independently} for the two experts,
covering the whole $(\tau_a,\tau_z)$ square; the network receives $t=(1-\tau)\cdot 1000$
through a $256$-dim sinusoidal embedding followed by a two-layer MLP, and the clean prefix is
conditioned with $t=0$ (i.e., $\tau=1$). Both terms are per-token MSE, averaged over the
feature axis, masked-averaged over valid steps/frames, then averaged over the batch, and
combined as $\mathcal{L}=\mathcal{L}_a+\lambda_z\mathcal{L}_z$ with $\lambda_z=1$.

\paragraph{Shared optimization protocol.} All configurations share: AdamW
($\beta=(0.9,0.95)$, weight decay $10^{-2}$), peak learning rate $2\times10^{-4}$, linear warm-up
over the first $5\%$ of the total steps then cosine decay to $1\%$ of the peak, batch size $32$ on
a single GPU with no gradient accumulation, bf16 mixed precision , seed $42$, no EMA, no dropout, and the same data pipeline
(Sec.~\ref{app:observation_target}): $16$-step action chunks, four future latents at
$t\!+\!4,\dots,t\!+\!16$, per-channel latent normalization, and $277{,}713$ training windows
($8{,}679$ optimizer steps per epoch). Only the visual encoder, the denoiser width, the epoch
budget and the gradient-clipping threshold differ across rows. Checkpoints are written every
$5{,}000$ steps and we always report the \emph{final} checkpoint; no checkpoint selection on
evaluation results is performed.

\begin{center}
\small
\begin{tabular}{llccccc}
\toprule
Encoder & Denoiser & $D_z$ & trainable & epochs & steps & grad.\ clip \\
\midrule
V-JEPA~2.1 ViT-B/16 (80M)  & DiT-S & 768  & 63.41M  & 20 & 173,580 & 1.0 \\
V-JEPA~2.1 ViT-L/16 (300M) & DiT-S & 1024 & 63.60M  & 20 & 173,580 & 1.0 \\
V-JEPA~2.1 ViT-L/16 (300M) & DiT-B & 1024 & 245.76M & 15 & 130,185 & 2.0 \\

\bottomrule
\end{tabular}
\end{center}

\section{Evaluation Protocols and Baseline Provenance}
\label{app:evaluation_protocols}

All evaluations are closed-loop in the original simulators with a single policy implementation.
No evaluation-time tuning, action ensembling, or test-time augmentation is used. Common
inference settings are listed in Table~\ref{tab:eval_settings}.

\begin{table}[h]
\centering
\small
\caption{Closed-loop inference settings.}
\label{tab:eval_settings}
\begin{tabular}{lccc}
\toprule
 & LIBERO & LIBERO-Plus & RoboTwin~2.0 \\
\midrule
Action chunk / executed steps & 16 / 10 & 16 / 10 & 32 / 24 \\
Euler steps                   & 4 & 4 & 10 \\
CFG scale                     & 1  & 1.5 & 1 \\
Future-image generation       & off & off & off \\
Warm-up no-op steps           & 30 & 30 & -- \\

\bottomrule
\end{tabular}
\end{table}

\subsection{LIBERO and LIBERO-Plus}
\label{app:libero_protocol}

\paragraph{Dataset and evaluation.} We evaluate on LIBERO and LIBERO-Plus. For LIBERO, we jointly
train on the four standard task suites (Spatial, Object, Goal, Long; $1{,}712$ demonstrations,
$277{,}713$ frames) and evaluate the resulting policy on the corresponding test tasks: $10$ tasks
per suite with $50$ trials each, i.e.\ $2{,}000$ episodes per checkpoint. Initial states are taken
from the benchmark's own init-state files indexed by trial number, so every model sees identical
start configurations, and the simulator is seeded with $42$. For LIBERO-Plus, we directly transfer
the policy trained on LIBERO demonstrations to the perturbed environments without additional
fine-tuning. LIBERO-Plus provides $10{,}030$ perturbed task instances over seven perturbation
dimensions (Camera, Robot, Language, Light, Background, Noise, Layout), each evaluated for a
single episode; we report the success rate per dimension and the instance-weighted overall success rate used by the LIBERO-Plus leaderboard. For ablations, we use a frozen stratified subsample
of 2,640 instances (660 per suite, balanced over the seven dimensions, drawn once with seed 42 and
reused by every run). Also, we report the results without CFG which yields a slightly lower success rates than the main table. The comparison fairness between different variants are always ensured.

\paragraph{Observation and action configuration.} The policy receives images from the primary
camera and the wrist camera together with language instructions and proprioceptive states. The two
views are rendered at $256\times256$, resized to $224\times224$ each and concatenated horizontally
into a $224\times448$ input, which the frozen V-JEPA encoder maps to $392$ visual tokens. Language
is encoded offline into $128$ tokens and proprioception is $8$-dimensional (end-effector pose and
two gripper joints). The action space contains $7$-dimensional robot actions ($6$ OSC pose deltas
and one gripper command) with an action chunk length of $16$.

\paragraph{Closed-loop execution.} Each episode begins with $30$ no-op steps so the scene settles.
The policy then denoises a chunk with $4$ Euler steps, executes the first $10$ actions, and
replans; an observation is taken after every simulator step so that the four-frame, stride-two
clip conditioning the policy keeps the temporal spacing used in training. Predicted actions are
de-normalized with the checkpoint's own dataset statistics and the gripper channel is binarized.
The episode budget is $400$ steps for Spatial, Object and Goal and $700$ steps for Long. Training
hyper-parameters are given in Sec.~\ref{app:training_details}.

\subsection{RoboTwin 2.0}
\label{app:robotwin_protocol}

\paragraph{Dataset and evaluation.} We evaluate on RoboTwin~2.0 using all $50$ manipulation tasks.
For each task the policy is trained on a fixed subset of $50$ demonstrations ($2{,}500$ episodes,
$564{,}911$ frames in total, selected once with seed $42$). The same trained policy is then
evaluated on both the Clean and the Random settings, where Random introduces randomized
backgrounds, a cluttered table, randomized lighting and table-height jitter. We run $25$ episodes
per task and setting ($2{,}500$ evaluation episodes) and report the average task success rate over
all $50$ tasks. Language instructions at evaluation are drawn from the unseen template pool, and
evaluation scenes are disjoint from training scenes by construction: episode seeds start at
$4{,}300{,}000$ whereas the demonstrations were collected from seeds starting at $0$. Per-episode
step limits and success criteria are the benchmark's own.

\paragraph{Observation and action configuration.} RoboTwin provides one external camera and two
wrist cameras at $240\times320$. The three views are composed into a single $384\times320$ canvas
(external view on top, the two wrist views side by side below) and encoded by the frozen V-JEPA
encoder into $480$ visual tokens. The policy predicts $14$-dimensional bimanual actions with an
action chunk length of $32$, of which $24$ steps are executed before replanning; denoising uses
$10$ Euler steps. Training follows Sec.~\ref{app:training_details} with a longer temporal stride
(clips of four frames at stride four, future latents at $+8,\dots,+32$) to match the $50$\,Hz
control rate.

\paragraph{Per-task results.} Complete task-wise results for all $50$ tasks are given in
Table~\ref{tab:robotwin_per_task}; the overall success rate is $57.8\%$ (Clean), $59.9\%$ (Random)
and $58.8\%$ on average.

\begin{table}[h]
\centering
\caption{RoboTwin~2.0 per-task success rate (\%), $25$ episodes per cell.
C = Clean, R = Random, M = mean.}
\label{tab:robotwin_per_task}
\tiny
\begin{tabular}{lccc|lccc}
\toprule
Task & C & R & M & Task & C & R & M \\
\midrule
adjust\_bottle & 100 & 100 & 100 & place\_can\_basket & 64 & 64 & 64 \\
beat\_block\_hammer & 76 & 64 & 70 & place\_cans\_plasticbox & 56 & 84 & 70 \\
blocks\_ranking\_rgb & 36 & 36 & 36 & place\_container\_plate & 100 & 96 & 98 \\
blocks\_ranking\_size & 24 & 32 & 28 & place\_dual\_shoes & 64 & 44 & 54 \\
click\_alarmclock & 92 & 84 & 88 & place\_empty\_cup & 72 & 72 & 72 \\
click\_bell & 100 & 92 & 96 & place\_fan & 40 & 56 & 48 \\
dump\_bin\_bigbin & 88 & 88 & 88 & place\_mouse\_pad & 28 & 20 & 24 \\
grab\_roller & 80 & 96 & 88 & place\_object\_basket & 52 & 60 & 56 \\
handover\_block & 64 & 36 & 50 & place\_object\_scale & 44 & 12 & 28 \\
handover\_mic & 92 & 96 & 94 & place\_object\_stand & 76 & 72 & 74 \\
hanging\_mug & 28 & 32 & 30 & place\_phone\_stand & 48 & 60 & 54 \\
lift\_pot & 44 & 60 & 52 & place\_shoe & 76 & 68 & 72 \\
move\_can\_pot & 48 & 68 & 58 & press\_stapler & 60 & 84 & 72 \\
move\_pillbottle\_pad & 44 & 32 & 38 & put\_bottles\_dustbin & 36 & 52 & 44 \\
move\_playingcard\_away & 64 & 68 & 66 & put\_object\_cabinet & 36 & 64 & 50 \\
move\_stapler\_pad & 8 & 8 & 8 & rotate\_qrcode & 72 & 68 & 70 \\
open\_laptop & 88 & 88 & 88 & scan\_object & 24 & 36 & 30 \\
open\_microwave & 20 & 24 & 22 & shake\_bottle & 100 & 100 & 100 \\
pick\_diverse\_bottles & 36 & 48 & 42 & shake\_bottle\_horizontally & 100 & 100 & 100 \\
pick\_dual\_bottles & 68 & 52 & 60 & stack\_blocks\_three & 8 & 28 & 18 \\
place\_a2b\_left & 36 & 36 & 36 & stack\_blocks\_two & 80 & 84 & 82 \\
place\_a2b\_right & 56 & 40 & 48 & stack\_bowls\_three & 64 & 60 & 62 \\
place\_bread\_basket & 36 & 44 & 40 & stack\_bowls\_two & 100 & 92 & 96 \\
place\_bread\_skillet & 60 & 52 & 56 & stamp\_seal & 16 & 28 & 22 \\
place\_burger\_fries & 56 & 84 & 70 & turn\_switch & 28 & 32 & 30 \\
\midrule
\multicolumn{4}{l|}{\textbf{Overall ($50$ tasks)}} & & \textbf{57.8} & \textbf{59.9} & \textbf{58.8} \\
\bottomrule
\end{tabular}
\end{table}

\section{Real-Robot Experiments}
\label{app:realworld}

We evaluate \model{} on two real-world platforms: a Unitree G1 humanoid for bimanual manipulation and a Fanuc CRX-10iA arm for tabletop manipulation. Unless otherwise specified, all experiments use the 300M V-JEPA encoder, DiT-S ($63.6$M trainable parameters), and four future latent targets at $+4,+8,+12,+16$ steps. We input four frames into V-JEPA and use the last latent frame as the DiT inputs. Optimization follows Sec.~\ref{app:training_details}.

\subsection{Unitree G1}
\label{app:g1_protocol}

\paragraph{Tasks and data.}
We consider three bimanual manipulation tasks on a Unitree G1 humanoid: \emph{Bimanual Plate Handover}, \emph{Box Packing}, and \emph{Cup Insertion}. Demonstrations are successful policy rollouts recorded at $30$,Hz and converted to LeRobot~v2.1 format. We train a separate policy for each task.

\paragraph{Observations and actions.}
The policy receives synchronized RGB observations from a third-person camera and a head-mounted egocentric camera. Both $640\times480$ images are center-cropped, resized to $224\times224$, and concatenated horizontally into a $224\times448$ input, yielding $392$ V-JEPA tokens. Proprioception and actions are $16$-dimensional absolute joint-position vectors, comprising $7$ joints per arm and two binary gripper states. Both are min--max normalized. The policy predicts $16$ actions per chunk, corresponding to $0.53$,s at $30$,Hz.

\begin{center}
\small
\begin{tabular}{lrrrrr}
\toprule
Task & Episodes & Frames & LR & Epochs & Steps \\
\midrule
Plate Handover & 167 & 71{,}333 & $2\times10^{-4}$ & 20 & 44{,}600 \\
Box Packing    & 185 & 47{,}859 & $2\times10^{-4}$ & 25 & 37{,}400 \\
Cup Insertion  & 170 & 61{,}814 & $2\times10^{-4}$ & 20 & 38{,}640 \\
\bottomrule
\end{tabular}
\end{center}

\paragraph{Deployment.}
At inference, observations are processed using the same image pipeline as during training. The instruction embedding is pre-computed and cached, so no text encoder is required on the robot. Actions are generated with $10$ Euler steps in action-only mode; future-image prediction is disabled at inference. Predicted chunks are de-normalized using the corresponding training-set statistics and executed in a receding-horizon manner at $30$ Hz. 

\subsection{Fanuc CRX-10iA}
\label{app:crx_protocol}

\paragraph{Tasks and data.}
We evaluate two tabletop manipulation tasks: \emph{Mug Hanging}, where the robot places a mug on a mug tree, and \emph{Multi-stage Cooking}, which requires opening a pot, placing corn into the pot, and transferring garlic onto a plate. Demonstrations are collected by human teleoperation using a $6$-DoF SpaceMouse at $10$ Hz and converted to LeRobot~v2.1. A separate model is trained for each task.

\paragraph{Observations and actions.}
A fixed third-person RealSense camera and a wrist-mounted RealSense camera provide two $240\times240$ RGB views. Each image is resized to $224\times224$ and the pair is concatenated into a $224\times448$ input ($392$ V-JEPA tokens). The action space is the $8$-dimensional SpaceMouse command

$$
[x,y,z,\mathrm{pitch},\mathrm{roll},\mathrm{yaw},\mathrm{btn}_1,\mathrm{btn}_2],
$$

while proprioception consists of the commanded Cartesian pose and gripper angle. Actions and states are min--max normalized. Because the action representation is incremental, padded action steps are set to zero. Each prediction contains $16$ actions, corresponding to $1.6$,s at $10$,Hz.

\begin{center}
\small
\begin{tabular}{lrrrrr}
\toprule
Task & Episodes & Frames & LR & Epochs & Steps \\
\midrule
Mug Hanging & 136 & 25{,}124 & $1\times10^{-4}$ & 35 & 27{,}510 \\
Cooking Sequence & 81 & 19{,}936 & $1\times10^{-4}$ & 50 & 31{,}150 \\
\bottomrule
\end{tabular}
\end{center}

\paragraph{Deployment.}
The policy operates at $10$Hz using $10$ Euler denoising steps in action-only mode. At each planning step, the first $8$ actions of the $16$-step prediction are executed before re-observation and replanning. Text embeddings and normalization statistics are loaded from the training checkpoint. Episodes are limited to $1000$ control steps, and checkpoints are validated by open-loop replay on held-out demonstrations before real-world deployment.

\end{document}